%% file: Main.tex
\documentclass[reqno]{amsart}

\usepackage{amsmath}
\usepackage{amssymb}
\usepackage{amsfonts}
\usepackage{amsthm}
\usepackage[foot]{amsaddr}

\usepackage{mathtools}
\mathtoolsset{%
}

\usepackage[utf8]{inputenc}
\usepackage[T1]{fontenc}

\usepackage{libertine}
\usepackage[libertine,libaltvw,cmintegrals,vvarbb]{newtxmath}

\usepackage{dsfont}

\usepackage[%
cal=cm,
]
{mathalfa}

\usepackage[labelfont={bf,small},labelsep=colon,font=small]{caption}
\usepackage[dvipsnames,svgnames]{xcolor}
\colorlet{MyBlue}{DodgerBlue!40!Black}
\colorlet{MyGreen}{DarkGreen!85!Black}

\usepackage{subfigure}
\usepackage{tikz}
\usetikzlibrary{calc,patterns}

\usepackage{acronym}
\usepackage{booktabs}
\usepackage{latexsym}
\usepackage{paralist}
\usepackage{wasysym}
\usepackage{xspace}

\usepackage[numbers,sort&compress]{natbib}

\usepackage{hyperref}
\hypersetup{
draft=false,
colorlinks=true,
linktocpage=true,
pdfstartview=FitH,
breaklinks=true,
pdfpagemode=UseNone,
pageanchor=true,
pdfpagemode=UseOutlines,
plainpages=false,
bookmarksnumbered,
bookmarksopen=false,
bookmarksopenlevel=1,
hypertexnames=true,
pdfhighlight=/O,
urlcolor=MyBlue,linkcolor=MyBlue,citecolor=MyGreen, 
pdftitle={},
pdfauthor={},
pdfsubject={},
pdfkeywords={},
pdfcreator={pdfLaTeX},
pdfproducer={LaTeX with hyperref}
}

\numberwithin{equation}{section}  
\usepackage[sort&compress,capitalize,nameinlink]{cleveref}
\crefname{app}{Appendix}{Appendices}

\crefrangeformat{equation}{\upshape(#3#1#4)\textendash(#5#2#6)}

\newcommand{\dd}{\:d}

\newcommand{\eps}{\varepsilon}
\newcommand{\from}{\colon}

\newcommand{\mg}{\succ}
\newcommand{\mgeq}{\succcurlyeq}

\newcommand{\simplex}{\Delta}

\newcommand{\bH}{\mathbf{H}}
\newcommand{\bI}{\mathbf{I}}

\newcommand{\bM}{\mathbf{M}}

\newcommand{\bQ}{\mathbf{Q}}

\newcommand{\bV}{\mathbf{V}}
\newcommand{\bW}{\mathbf{W}}
\newcommand{\bX}{\mathbf{X}}
\newcommand{\bY}{\mathbf{Y}}

\newcommand{\bp}{\mathbf{p}}
\newcommand{\bq}{\mathbf{q}}

\newcommand{\R}{\mathbb{R}}

\DeclareMathOperator*{\argmax}{arg\,max}
\DeclareMathOperator*{\argmin}{arg\,min}

\DeclareMathOperator{\bigoh}{\mathcal O}

\DeclareMathOperator{\diam}{diam}

\DeclareMathOperator{\eig}{eig}

\DeclareMathOperator{\ex}{\mathbb{E}}
\DeclareMathOperator{\Hess}{Hess}

\DeclareMathOperator{\one}{\mathds{1}}

\DeclareMathOperator{\prob}{\mathbb{P}}

\DeclareMathOperator{\tr}{tr}

\DeclareMathOperator{\vol}{vol}

\providecommand\given{} 

\DeclarePairedDelimiter{\braces}{\{}{\}}
\DeclarePairedDelimiter{\bracks}{[}{]}
\DeclarePairedDelimiter{\parens}{(}{)}

\DeclarePairedDelimiter{\abs}{\lvert}{\rvert}
\DeclarePairedDelimiter{\norm}{\lVert}{\rVert}

\DeclarePairedDelimiter{\ceil}{\lceil}{\rceil}

\DeclarePairedDelimiterX{\braket}[2]{\langle}{\rangle}{#1,#2}

\newcommand{\product}[2]{#1^{\top}#2}
\DeclarePairedDelimiterX{\setdef}[2]{\{}{\}}{#1:#2}

\DeclarePairedDelimiterXPP{\exclude}[1]{\mathopen{}\setminus}{\{}{\}}{}{#1}

\DeclarePairedDelimiterXPP{\probof}[1]{\prob}{(}{)}{}{%
\renewcommand\given{\nonscript\:\delimsize\vert\nonscript\:\mathopen{}}
#1}

\DeclarePairedDelimiterXPP{\exof}[1]{\ex}{[}{]}{}{%
\renewcommand\given{\nonscript\:\delimsize\vert\nonscript\:\mathopen{}}
#1}

\DeclarePairedDelimiterXPP{\trof}[1]{\tr}{[}{]}{}{#1}

\newcommand{\txs}{\textstyle}
\newcommand{\textpar}[1]{\textup(#1\textup)}

\usepackage[textwidth=30mm]{todonotes}
\newcommand{\debug}[1]{#1}

\newcommand{\hilite}[1]{\emph{#1}}

\usepackage{algorithm}
\usepackage{algorithmic}

\makeatletter
\newenvironment{process}[1][H]{%
    \renewcommand{\ALG@name}{Process} 
   \begin{algorithm}[#1]%
  }{\end{algorithm}}
\makeatother

\theoremstyle{plain}
\newtheorem{theorem}{Theorem}

\newtheorem*{corollary*}{Corollary}

\theoremstyle{definition}

\newtheorem*{definition*}{Definition}

\theoremstyle{remark}

\newtheorem*{remark*}{Remark}
\newtheorem{example}{Example}
\newtheorem*{example*}{Example}

\numberwithin{theorem}{section}
\numberwithin{remark}{section}
\numberwithin{example}{section}

\newcommand{\unitvec}{\debug z}

\newcommand{\vdim}{\debug d}

\newcommand{\feas}{\mathcal{\debug X}}

\newcommand{\base}{\debug p}

\newcommand{\test}{\base}

\newcommand{\obj}{\debug f}
\newcommand{\Lip}{\debug L}

\newcommand{\diamfeas}{\diam(\feas)}
\newcommand{\ball}{\mathbb{B}}
\newcommand{\sphere}{\mathbb{S}}

\newcommand{\strong}{\debug \beta}
\newcommand{\breg}{\debug D}
\newcommand{\prox}{\debug P}

\DeclareMathOperator{\Eucl}{\debug \Pi}
\DeclareMathOperator{\logit}{\debug \Lambda}

\newcommand{\pure}{\debug a}

\newcommand{\nPures}{\debug A}
\newcommand{\pures}{\mathcal{\nPures}}

\newcommand{\pay}{\debug u}
\newcommand{\payv}{\debug v}
\newcommand{\loss}{\debug \ell}

\newcommand{\reg}{\debug R}
\newcommand{\dreg}{\reg^{\ast}}
\newcommand{\gap}{\debug \Delta}

\newcommand{\mean}{\debug \mu}
\newcommand{\sobj}{\debug F}
\newcommand{\sample}{\debug \omega}

\newcommand{\filter}{\mathcal{F}}

\newcommand{\est}{\hat\payv}

\newcommand{\opt}[1]{#1^{\ast}}

\newcommand{\step}{\debug \gamma}
\newcommand{\mix}{\debug \delta}

\newcommand{\act}{\debug x}
\newcommand{\actalt}{\act'}
\newcommand{\score}{\debug y}
\newcommand{\pivot}{\hat\act}

\newcommand{\vbound}{\debug V}

\newcommand{\channel}{\pure}
\newcommand{\nChannels}{\nPures}
\newcommand{\channels}{\mathcal{\nChannels}}

\newcommand{\rate}{C}

\newcommand{\pmax}{P_{\!\max}}
\newcommand{\rx}{N}
\newcommand{\tx}{M}

\newcommand{\iRun}{\debug s}
\newcommand{\run}{\debug t}
\newcommand{\horizon}{\debug T}
\newcommand{\nTimes}{\debug n}
\newcommand{\window}{\debug W}

\newcommand{\acdef}[1]{\emph{\acl{#1}} \textup(\acs{#1}\textup)\acused{#1}}
\newcommand{\acdefp}[1]{\emph{\aclp{#1}} \textup(\acsp{#1}\textup)\acused{#1}}

\newcommand{\eg}{e.g.,\xspace}
\newcommand{\ie}{i.e.,\xspace}
\newcommand{\cf}{cf.\xspace}

\newacro{MDML}{mirror descent for metric learning}
\newacro{SSA}{simultaneous stochastic approximation}
\newacro{EG}{exponentiated gradient}
\newacro{MD}{mirror descent}
\newacro{EPA}{extended pedestrian A}
\newacro{EVA}{extended vehicular A}
\newacro{ETU}{extended typical urban}
\newacro{INF}{implicitly normalized forecaster}
\newacro{KL}{Kullback\textendash Leibler}
\newacro{OFDMA}{orthogonal frequency-division multiple access}
\newacro{MXL}{matrix exponential learning}
\newacro{IWF}{iterative water-filling}
\newacro{SWF}{simultaneous water-filling}
\newacro{OMD- bla}{bla}
\newacro{OMD-EG}{normalized exponentiated gradient}
\newacro{MAB}{multi-armed bandit}
\newacro{MIMO}{multiple-input and multiple-output}
\newacro{MU}{multi-user}
\newacro{EW}{exponential weights}
\newacro{MW}{multiplicative weights}
\newacro{EXP3}{exponential-weight algorithm for exploration and exploitation}
\newacro{OGA}{online gradient ascent}
\newacro{OGD}{online gradient descent}
\newacro{OGD-0}{online gradient descent with zeroth-order feedback}
\newacro{OCO}{online convex optimization}
\newacro{OSCO}{online strongly convex optimization}
\newacro{OMD}{online mirror descent}
\newacro{UCB}{upper confidence bound}
\newacro{SP}{signal processing}
\newacro{OLP}{online learning process}
\newacro{iid}[i.i.d.]{independent and identically distributed}
\newacro{LHS}{left-hand side}
\newacro{RHS}{right-hand side}

\begin{document}

\title
[Online Convex Optimization and No-Regret Learning]
{Online Convex Optimization and No-Regret Learning:\\
Algorithms, Guarantees and Applications}

\author
[E.~V.~Belmega, P.~Mertikopoulos, R.~Negrel, and L.~Sanguinetti]
{E.~Veronica Belmega$^{\ast}$
\and
Panayotis Mertikopoulos$^{\S}$
\and\\
Romain Negrel$^{\ddag}$
\and
Luca Sanguinetti$^{\P}$}

\address{$^{\ast}$ ETIS, UMR 8051, Université Paris Seine, Université Cergy-Pontoise, ENSEA, CNRS, France}
\address{$^{\S}$ Univ. Grenoble Alpes, CNRS, Inria, LIG, F-38000 Grenoble, France.}
\address{$^{\ddag}$ LIGM/ESIEE Paris-Université Paris-Est, Marne-la-Vallée, France.}
\address{$^{\P}$ University of Pisa, Dipartimento di Ingegneria dell'Informazione, Italy, and
Networks Group (LANEAS), CentraleSupélec, Université Paris-Saclay, Gif-sur-Yvette, France}

%
%
%

\thanks{This work was supported by
the French National Research Agency (ANR) project ORACLESS,
the Huawei Flagship project ULTRON,
and by ENSEA, Cergy-Pontoise, France.}


\begin{abstract}
\input{Abstract}
\end{abstract}

\allowdisplaybreaks
\acresetall
\acused{iid}
\maketitle

\section{Introduction}
\label{sec:intro}
\input{Introduction}

\section{A Gentle Start: Multi-Armed Bandits}
\label{sec:bandits}
\input{Bandits}

\section{Online Optimization}
\label{sec:setting}
\input{Setting}

\section{Online Algorithms and Guarantees}
\label{sec:algorithms}
\input{Algorithms}

\section{Additional Considerations}
\label{sec:extras}
\input{Extras}

\section{Concluding remarks}
\label{sec:conclusions}
\input{Conclusions}

\bibliographystyle{siam}
\bibliography{IEEEabrv,Bibliography-SPM}

\end{document}

%% file: Abstract.tex
%
%
Spurred by the enthusiasm surrounding the ``Big Data'' paradigm, the mathematical and algorithmic tools of online optimization have found widespread use in problems where the trade-off between data exploration and exploitation plays a predominant role.
This trade-off is of particular importance to several branches and applications of signal processing, such as data mining, statistical inference, multimedia indexing and wireless communications (to name but a few).
With this in mind, the aim of this tutorial paper is to provide a gentle introduction to online optimization and learning algorithms that are asymptotically optimal in hindsight \textendash\ \ie they approach the performance of a virtual algorithm with unlimited computational power and full knowledge of the future,
a property known as \emph{no-regret}.
Particular attention is devoted to identifying the algorithms' theoretical performance guarantees and to establish links with classic optimization paradigms (both static and stochastic).
To allow a better understanding of this toolbox, we provide several examples throughout the tutorial ranging from metric learning to wireless resource allocation problems.

%% file: Introduction.tex

The prototypical setting of online optimization is as follows:
at each stage of a repeated decision process, an optimizing agent makes a prediction and incurs a loss based on an a priori \emph{unknown} cost function that measures the accuracy of said prediction.
Subsequently, the agent receives some problem-specific feedback and makes a new prediction, possibly against a different ground truth (empirical evidence), with the goal of improving the quality of their predictions over time.
In this broad context, the theory of online optimization seeks to characterize the limits of what is achievable in terms of performance guarantees, and to provide efficient algorithms to achieve them.

Spurred by the enthusiasm surrounding the Big Data paradigm, the mathematical and algorithmic tools developed in the field of online optimization have found widespread use in problems where the trade-off between data \emph{exploration} and \emph{exploitation} plays a predominant role.
In signal processing, examples of this include
sparse coding and dictionary learning \cite{MBPS10},
data classification and filtering \cite{GH13},
matrix completion and prediction \cite{SSS14},
and many others.
More recently, motivated by the highly dynamic nature of future and emerging wireless networks, online optimization methods have been successfully exploited to design resource allocation policies for problems ranging from signal covariance optimization in multi-antenna terminals \cite{MB16} to channel selection and cognitive medium access \cite{AMTS11}.

\begin{figure}[t]
\centering
\input{Figures/Framework.tex}
\caption{A bird's eye view of the links between online optimization and other optimization paradigms.}
\label{fig:framework}
\end{figure}
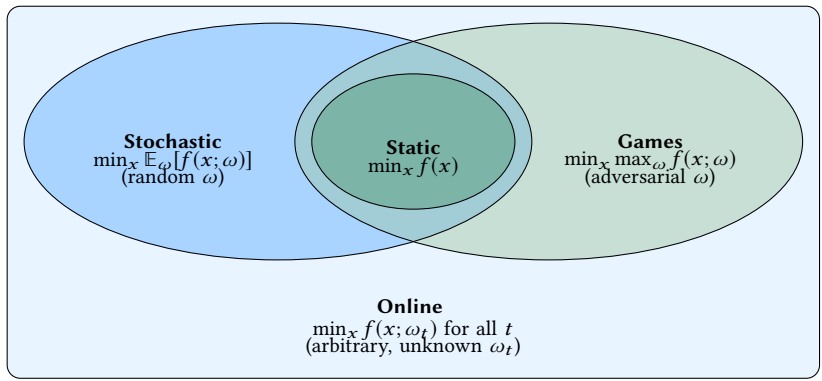

In classic, static optimization problems, the core underlying assumption is that the objective to be optimized is known by the optimizing agent and remains fixed for the entire runtime of the algorithm computing a solution.
Stochastic optimization provides an extension of this framework to problems where the objective function may also depend on a stationary stochastic process.
Game theory takes an alternative, multi-agent view of such problems, often revolving around worst-case guarantees against an adversary \textendash\ \eg as in von Neumann's celebrated minimax vision of zero-sum games.
However, all these extensions rely on strong assumptions regarding the variability of the problem's objective, the agents' rationality and common knowledge of rationality (in games), the information at the agents' disposal, etc.
By contrast, online optimization provides an elegant toolbox which goes beyond the above by allowing for variations in the problem that are \emph{completely arbitrary} \textendash\ typically accounting for exogenous (stationary or otherwise) parameters affecting it (cf.~Fig.~\ref{fig:framework}).


A key example of this framework is encountered in wireless communication networks, where traditional static/stationary paradigms no longer suffice to capture and cope with issues such as
non-random fluctuations in the wireless medium,
the unpredictable behavior of the communicating devices in distributed networks, etc. \cite{MB14,MB16}.
Likewise, in metric learning problems for multimedia indexing and classification, the online framework allows to exploit distributed annotations from multiple input sources (such as the Internet users), as opposed to relying on centralized annotation of very large data sets, and also to relax all the stochastic assumptions among annotations that are common in offline metric learning \cite{Kunapuli2012}.

In addition to its wide-ranging scope, another major advantage of the online framework is that the derived algorithms \textendash\ referred here as \emph{online policies} \textendash\ come with \hilite{provable theoretical guarantees in the face of uncertainty}.
These guarantees are based on Hannan's definition of \hilite{regret} \cite{Han57}, a seminal notion which compares an algorithm's performance to that of the best fixed policy in hindsight, an idealized benchmark that minimizes the total loss incurred over the horizon of play with perfect knowledge of the future.
Since this benchmark requires full, non-causal knowledge of the dynamics governing the evolution of the problem's objective, this is obviously not an implementable policy \textendash\ it only exists as a theoretical performance target.
As such, the aim of online optimization is to derive causal, online learning algorithms that get as close as possible to this formidable target with the fewest possible assumptions.
What is truly striking is that it succeeds:
the last two decades have seen a flurry of activity on the design of online algorithms that attain the best possible regret minimization rates in a broad range of problems, thus explaining the widespread use of such algorithms in Big Data.

We dive in all this by first exploring in \cref{sec:bandits} a fundamental \emph{discrete choice} problem that has been the focus of a very active and vigorous literature:
\acdefp{MAB}.
This problem is simple to state and provides a gentle introduction to the fundamental exploration\textendash exploitation trade-off that underlies much of online optimization;
at the same time, it is sufficiently deep to require highly sophisticated analytical tools to solve it.
A key lesson from this problem is that deterministic online policies suffice to achieve no regret in (stationary) stochastic environments.
However, when pitted against an informed \emph{adversary}, such algorithms are doomed to fail:
when ``gambling in a rigged casino'', only randomized algorithms can surprise the adversary often enough to attain a no-regret state.

Extending the above to problems with continuous action spaces, we introduce in \cref{sec:setting} the fundamentals of online optimization (along with several applications and examples), and we discuss in \cref{sec:algorithms} the most widely studied classes of online algorithms and their theoretical guarantees (again, with several practical applications).
Specifically, we focus on Zinkevich's seminal \acdef{OGD} method \cite{Zin03} and on its generalization: the class of \acdef{OMD} schemes pioneered by Shalev-Shwartz \cite{SS07,SS11}.
By tailoring each method to the geometry of the problem at hand, it is possible to attain regret minimization rates that are provably optimal in terms of the problem's dimension, going as far as achieving a logarithmic reduction thereof.
This is a powerful instance of lifting the \hilite{curse of dimensionality:}
in many machine learning and Big Data applications, the dimension of the problem's state space is exponentially large, so these techniques render the solution of such problems feasible.
Finally, we discuss in \cref{sec:extras} the impact of noise and imperfections in the feedback received by the optimizer, the possibility to learn without gradient-based feedback, and other considerations that arise in practice.

To sum up, this tutorial is intended for researchers and graduate students interested in an overview of online optimization and no-regret learning, including theoretical guarantees, the algorithms that achieve them, and their applications.
More specifically, our goal is to:
\begin{itemize}
\item
Provide a holistic overview of the topic highlighting its connections with other optimization paradigms.
\item
Delineate the fundamental theoretical limits of what can be achieved and present the algorithms that achieve these limits.
\item
Explain to what extent the feedback limitations that arise in practical signal processing applications affect the performance of such algorithms.
\item
Provide applications and examples from different areas of signal processing to help the reader appreciate the wide applicability of online optimization and no-regret learning methods.
\end{itemize}

Our goal is to give an introduction to these questions, without going into overwhelming detail.
In particular, we make no effort or claim of providing a comprehensive survey of the state-of-the-art;
this would require a much more in-depth treatment than an introductory tutorial is meant to provide, so we do not attempt it.

\subsubsection*{Landau notation}
Throughout this tutorial, we will make heavy use of the Landau asymptotic notation $\bigoh(\cdot)$, $o(\cdot)$, $\Omega(\cdot)$, etc.
First, given two functions $f,g\from\R\to\R$, we say that ``$f$ grows no faster than $g$ asymptotically'' and we write $f(\run)=\bigoh(g(\run))$ as $\run\to\infty$ if $\limsup_{\run\to\infty} f(\run)/g(\run) < \infty$, \ie there exists some positive constant $c>0$ such that $\abs{f(\run)} < g(\run)$ for sufficiently large $\run$.
Conversely, we say that ``$f$ grows no slower than $g$'' and we write $f(\run) = \Omega(g(\run))$ if $g(\run) = \bigoh(f(\run))$ or, equivalently, $\limsup_{\run\to\infty} g(\run)/f(\run) > 0$.
If we have both $f(\run) = \bigoh(g(\run))$ and $f(\run) = \Omega(g(\run))$, we say that ``$f$ grows as $g$'' and we write $f(\run)=\Theta(g(\run))$.
Finally, if $\limsup_{\run\to\infty} f(\run)/g(\run) \leq 0$, we write $f(\run) = o(g(\run))$ and we say that ``$f$ is dominated by $g$'' (we ignore the negative part).

%% file: Figures/Framework.tex

\colorlet{BackgroundColor}{DodgerBlue!40!MidnightBlue}

\footnotesize
\begin{tikzpicture}
[scale=.9,
>=stealth,
vecstyle/.style = {->, line width=.5pt},
edgestyle/.style={-, line width=.5pt},
nodestyle/.style = {circle,fill=black,inner sep=0, minimum size=2},
plotstyle/.style={color=DarkGreen!80!Cyan,thick}]

\coordinate (static) at (0,0);
\coordinate (stochastic) at (-2,0);
\coordinate (games) at (2,0);

\filldraw [fill = DodgerBlue!10!white, rounded corners = 5] (-6,-3.5) rectangle (6,2);

\def\static{(static) ellipse [x radius = 1.5, y radius = 1]}
\def\stochastic{(stochastic) ellipse [x radius = 3.75, y radius = 1.75]}
\def\games{(games) ellipse [x radius = 3.75, y radius = 1.75]}

\fill [DarkGreen!50] \static;
\fill [DodgerBlue!80, fill opacity = .4] \stochastic;
\fill [DarkGreen!40, fill opacity = .4] \games;

\draw \static;
\draw \stochastic;
\draw \games;

\node (online) [text width = 5cm, align=center]
at (0,-2.75)
{%
\textsf{\bfseries Online}
\smallskip
\\[-1ex]
$\min_{\act} \obj(\act;\sample_{\run})$ for all $\run$
\\[-1ex]
(arbitrary, unknown $\sample_{\run}$)};

\node (static) [text width = 3 cm, align = center]
at (0,-.25)
{%
\textsf{\bfseries Static}
\\[-1ex]
$\min_{\act} \obj(\act)$};

\node (stochastic) [text width = 3.5cm, align=center]
at (-3.55,-1em)
{%
\textsf{\bfseries Stochastic}
\\[-1ex]
$\min_{\act} \ex_{\sample}[\obj(\act;\sample)]$
\\[-1ex]
(random $\sample$)};

\node (games) [text width = 3.5cm, align = center]
at (3.5,-1em)
{%
\textsf{\bfseries Games}
\\[-1ex]
$\min_{\act} \max_{\sample} \obj(\act;\sample)$
\\[-1ex]
(adversarial $\sample$)};

\end{tikzpicture}

%% file: Bandits.tex

We begin with the \acdef{MAB} problem, a fundamental \emph{sequence prediction} (or \emph{forecasting}) problem that permeates many areas of online and machine learning, and which has found a remarkable breadth of applications \textendash\ 
from sparse coding and dictionary learning \cite{MBPS10},
to
filtering \cite{GH13},
matrix prediction \cite{SSS14},
channel selection \cite{AMTS11},
antenna beam selection in mmWave communications \cite{Hashemi-2017}
and many others.
In a certain sense, \aclp{MAB} comprise the ``discrete'' analogue of the more general online optimization framework that we treat in the rest of this monograph.
We begin with this case mostly for historical reasons \textendash\ but also to illustrate some of the key notions involved, which is easier to do in the \ac{MAB} setting.

After a brief overview, we introduce the notion of regret in the context of stochastic \aclp{MAB} and we present the \acdef{UCB} algorithm, a deterministic policy that achieves no regret in this problem.
Beyond stochastic bandits, when facing an adversary (sometimes referred to as ``gambling in a rigged casino''), deterministic policies are no longer able to attain a no-regret state, so we turn to randomized strategies and present the regret guarantees of the \acdef{EW} algorithm, with both full and partial information.

\subsection{Problem Statement}
\label{sec:bandits-statement}

Dating back to the seminal work of Thompson \cite{Tho33} and Robbins \cite{Rob52}, a \acl{MAB} problem can be described as follows:
At each stage $\run=1,2,\dotsc$, of a repeated decision process,
an optimizing \emph{agent} (the decision-maker) selects an \emph{action} $\pure_{\run}$ from some finite set $\pures = \{1,\dotsc,\nPures\}$.
Based on this choice, the agent receives a \emph{reward} (or \emph{payoff}) $\pay_{\run}(\pure_{\run})$, they select a new action $\pure_{\run+1}$ and the process repeats.
Sometimes it is more convenient to state the problem in terms of incurred \emph{losses} instead of received payoffs;
when this is the case, we will write $\loss_{\run}(\pure_{\run}) = -\pay_{\run}(\pure_{\run})$ for the loss incurred by the agent at stage $\run$.

Of course, the interpretation of the agents' actions and their rewards (or losses) is context-specific and depends on the application at hand.
In Robbins' original formulation, the agent was a gambler choosing a slot machine in a casino (a ``one-armed bandit'' in the colorful slang of the 50's), and the agent's reward was the amount of money received minus the cost of playing \cite{Rob52}.
In clinical trials (the setting of Thompson's work), the choice of action represents the drug administered to a test patient and the incurred loss is the patient's time to recovery \cite{Tho33}.
A concrete example from wireless communications is as follows:

\begin{example}[Channel selection]
Referring to \cite{AMTS11} for the details, consider a wireless user transmitting over a set of $\channels = \{1,\dotsc,\nChannels\}$ of non-overlapping frequency channels.
At each time slot $\run=1,2,\dotsc$, a channel $\channel\in\channels$ could be either ``free'' or ``busy'':
if the chosen channel is free, the user's packet is transmitted successfully, otherwise there is a collision and the packet is lost.
Writing $\payv_{\channel,\run} \in\{0,1\}$ for the state of channel $\channel$ at time $\run$ (with $\payv_{\channel,\run}=0$ indicating that the channel is free), the user's loss at time $\run$ is $\loss_{\run}(\channel_{\run}) = -\payv_{\channel_{\run},\run}$, indicating a loss of $1$ whenever a packet drop occurs.
Hence, to maximize throughput, the user's objective is to minimize the total loss $\sum_{\run=1}^{\horizon} \loss_{\run}(\channel_{\run})$, \ie the total number of packets dropped over the horizon $\horizon$.
Since the channel statistics change with the number of users in the system (and are typically unknown to the users anyway), the main challenge is to design channel selection algorithms that minimize this loss ``on the fly'', \ie as information becomes available over time.
\end{example}

\paragraph*{Exploration vs. exploitation}
Returning to the general bandit setting, it is easy to see that the optimizer faces a trade-off between \hilite{exploration and exploitation.}
On the one hand, by trying out more arms (\ie ``\emph{exploring}''), the agent obtains more information and can make better choices in the future.
On the other hand, in so doing, the agent fails to take advantage of arms that yield better payoffs now, thus lagging behind in terms of performance (\ie ``\emph{exploitation}'').
As an example, in the channel selection problem, the more data used to train a scheduling algorithm, the better the channel selection will (eventually) become;
however, this comes at the expense of false positives (or negatives) that severely hamper the user's overall throughput.
Achieving \textendash\ and maintaining \textendash\ a balance between exploration and exploitation is the main objective of the literature on \aclp{MAB}.

Building on the above, there is a coarse categorization of \aclp{MAB} based on the way the sequence of payoff functions $\pay_{\run}\from\pures\to\R$ is generated.
These are as follows:

\begin{enumerate}
\setlength{\itemsep}{\smallskipamount}

\item
\emph{Stochastic bandits:}
Each arm $\pure\in\pures$ is associated to a statistical distribution $P_{\pure}$
and, at each stage $\run=1,2,\dotsc$,
the reward $\pay_{\run}(\pure)$ of the $\pure$-th arm is an \ac{iid} random variable $\payv_{\pure,\run}$ drawn from $P_{\pure}$ (\ie $\pay_{\run}(\pure) \equiv \payv_{\pure,\run} \sim P_{\pure}$ for all $\pure\in\pures$ and all $\run=1,2,\dotsc$).
The arms' reward distributions are not known to the agent, so the objective is to identify \textendash\ and \emph{exploit} \textendash\ the arm with the highest mean reward in as few trials as possible.

\item
\emph{Adversarial bandits:} The payoff of each arm is determined by an ``adversary'' (real or fictitious) who prescribes the $\run$-th stage ``payoff vector'' $\payv_{\run} = (\pay_{\run}(\pure))_{\pure\in\pures}$ at the same time that the agent chooses an arm.
In a certain sense, this is one of the most general formulations of the \ac{MAB} problem as all distributional assumptions are removed.
Thus, solving the adversarial problem also induces a solution to the stochastic bandit problem (though, perhaps, suboptimal).

\item
\emph{Markovian bandits:} The loss function $\pay_{\run}$ is determined by the state $\sample_{\run}$ of an underlying \textendash\ and possibly hidden \textendash\ Markov chain.
The transition matrix of this Markov chain is a priori unknown, and the optimizer's objective is to maximize their (discounted) expected reward over time.
\end{enumerate}

In what follows, we describe the stochastic and adversarial \ac{MAB} frameworks in detail.
Albeit historically significant, the discounted reward and information structure of Markovian bandits makes them less suitable for \acl{SP} applications, so they are not considered further in this tutorial;
the interested reader can find a recent survey of the topic in \cite{GGW11}.

\subsection{Stochastic Bandits}
\label{sec:bandits-stoch}

In a stochastic \acl{MAB}, the agent's objective would be to play the arm with the highest mean reward as many times as possible over the horizon of play $T$.
To quantify this, let $\mean_{\pure}$ denote the mean value of the reward distribution $P_{\pure}$ of arm $\pure\in\pures$ and let
\begin{equation}
\opt\mean
	= \max_{\pure\in\pures} \mean_{\pure}
	\quad
	\text{and}
	\quad
\opt\pure
	= \argmax_{\pure\in\pures} \mean_{\pure}
\end{equation}
respectively denote the bandit's maximal mean reward and the arm that achieves it (a priori, there could be several such arms but, for simplicity, we assume here that there is only one).
A standard assumption is to further posit that the bandits' rewards are always drawn from $[0,1]$:
this is simply a normalization convention motivated by the \emph{Bernoulli bandit} problem where each arm $\pure\in\pures$ returns a payoff of $1$ with probability $p_{\pure}$ and $0$ with probability $1-p_{\pure}$.%
\footnote{In the early literature on inference and stochastic bandits, the sequence prediction problem of identifying the probability that a Bernoulli process returns $1$ in as few trials as possible was synonymously referred to as a \emph{one-armed bandit}.}

\subsubsection{Regret}

With all this in mind, the agent's performance after $\horizon$ rounds can be quantified by aggregating the mean payoff gap between the bandit's best arm and the arm $\pure_{\run}$ chosen by the optimizer at each round $\run=1,2,\dots,\horizon$.
This quantity is known as the agent's mean \emph{regret} and is formally defined as%
\footnote{In the literature on \aclp{MAB}, $\bar\reg_{\horizon}$ is sometimes referred to as the pseudo-regret or weak regret \cite{CBL06,BCB12} to differentiate it from the expectation of the highest possible cumulative payoff generated by any given arm.
By the law of large numbers, the difference between the two is typically of the order of $\bigoh(\horizon^{1/2})$ in the case of independent draws, so the distinction disappears in settings where the worst possible regret scales as $\horizon^{1/2}$.}
\begin{equation}
\label{eq:reg-mean}
\bar\reg_{\horizon}
	= \sum_{\run=1}^{\horizon} \exof{\pay_{\run}(\opt\pure) - \pay_{\run}(\pure_{\run})}
	= \horizon\opt\mean - \sum_{\run=1}^{\horizon} \exof{\pay_{\run}(\pure_{\run})}.
\end{equation}
For an alternative interpretation of $\bar\reg_{\horizon}$,
let $\nTimes_{\pure,\run}$ denote the number of times that $\pure$ has been chosen up to time $\run$
and write $\gap_{\pure} = \opt\mean - \mean_{\pure}$ for the mean payoff difference between arm $\pure$ and the bandit's best arm.
Then, an easy rearrangement of \eqref{eq:reg-mean} yields the key expression
\begin{equation}
\label{eq:reg-mistakes}
\bar\reg_{\horizon}
	= \exof*{\sum_{\pure\in\pures} \nTimes_{\pure,\horizon}} \, \opt\mean
	- \exof*{\sum_{\pure\in\pures} \mean_{\pure} \nTimes_{\pure,\horizon}}
	= \sum_{\pure\in\pures} \gap_{\pure} \exof{\nTimes_{\pure,\horizon}}.
\end{equation}
Written this way, $\bar\reg_{\horizon}$ can be viewed as the mean number of suboptimal choices made by the optimizer,
weighted by the suboptimality gap $\gap_{\pure}$ of each arm;
put differently, $\bar\reg_{\horizon}$ can be seen as a measure of
\emph{the average number of mistakes made by the optimizer up to stage $\horizon$.}

In view of the above, maximizing the aggregate reward over $\horizon$ stages of play amounts to asking \emph{the regret be sublinear in $\horizon$}.
This is formally defined as
\begin{equation}
\label{eq:no-reg-mean}
\bar\reg_{\horizon}
	= o(\horizon),
\end{equation}
a property which is known as \hilite{no regret} and which guarantees that the relative number of mistakes (\ie draws of suboptimal arms) must vanish in the long run. 
As a result, much of the literature on stochastic bandits has focused on the design of algorithms that attain the best possible regret minimization rate in \eqref{eq:no-reg-mean}.

\subsubsection{No-regret algorithms for stochastic bandits}

A straightforward idea to optimize one's gains would be to keep a running average of the rewards obtained by each arm and then play the arm with the best estimate.
This \emph{pure exploration} (or ``follow the leader'') policy can be formally described as
\begin{equation}
\pure_{\run+1}
	= \argmax_{\pure\in\pures} \hat\mean_{\pure,\run}
\end{equation}
where
\begin{equation}
\hat\mean_{\pure,\run}
	= \frac{1}{\nTimes_{\pure,\run}} \sum_{\iRun=1}^{\run} \one(\pure_{\iRun}=\pure) \, \payv_{\pure,\iRun}
\end{equation}
denotes the empirical mean payoff of arm $\pure$ (\ie the total reward obtained from arm $\pure$ normalized by the number of times it was drawn).
This policy is a reasonable first try but it can easily get stuck at a suboptimal arm:
if the best arm performs very badly in its first draws, ``following the leader'' means it won't be drawn again in the future, so the agent's regret will grow linearly over time.

The above highlights the need for adding at least \emph{some} exploration into the mix:
otherwise, pure exploitation cannot lead to sublinear regret.
Building on this observation, the landmark idea of  \cite{LR85,ACBF02} was to \emph{retain optimism in the face of uncertainty,}
\ie to construct an \emph{``optimistic''} estimate for the mean payoff of each arm, and then pick the arm with the highest such estimate.
If this guess is wrong, optimism will fade quickly;
but if the guess is right, the agent will be able to balance exploration and exploitation, incurring little regret in the process.

The key ingredient of this ``optimism'' principle is provided by \emph{Hoeffding's inequality},
a fundamental tail bound in statistics which states that the sample mean $\bar X_{\nTimes} = \nTimes^{-1} \sum\nolimits_{i=1}^{\nTimes} X_{i}$ of a sequence of \ac{iid} random variables $X_{i}\in[0,1]$ with mean $\mean$ satisfies
\begin{equation}
\label{eq:Hoeffding}
\probof*{\bar X_{\nTimes} < \mean - z}
	\leq e^{-2\nTimes z^{2}}.
\end{equation}
In our bandit setting, if we take $z=\sqrt{\alpha\log \run/(2\nTimes_{\pure,\run})}$ for some $\alpha>0$, it follows that
\begin{equation}
\probof*{\hat\mean_{\pure,\run} + \sqrt{\alpha\log\run/(2\nTimes_{\pure,\run})} < \mean_{\pure}}
	\leq \run^{-\alpha},
\end{equation}
\ie the probability that the optimistic estimate $\hat\mean_{\pure,\run} + \sqrt{\alpha\log\run/(2\nTimes_{\pure,\run})}$ is less than $\mean_{\pure}$ decays very fast as a function of $\run$.


\begin{algorithm}[tbp]
\caption{\acf{UCB} with parameter $\alpha$}
\label{alg:UCB}
\input{Algorithms/UCB}
\end{algorithm}


Motivated by this, the \acdef{UCB} algorithm with tuning parameter $\alpha>2$ is defined via the recursion
\begin{equation}
\label{eq:UCB}
\pure_{\run+1}
	= \argmax_{\pure\in\pures} \braces*{\hat\mean_{\pure,\run} + \sqrt{\frac{\alpha\log\run}{2\nTimes_{\pure,\run}}}}.
\end{equation}
Heuristically, the first term in \eqref{eq:UCB} drives the agent to exploit the arm with the highest empirical mean while the second one encourages exploration by giving a second chance to arms which have not been played often enough (\ie $\nTimes_{\pure,\run}$ is small relative to $\run$).
This idea dates back to the landmark paper of Lai and Robbins \cite{LR85} and was subsequently expanded
by Agrawal \cite{Agr95},
Auer et al. \cite{ACBF02}, and many others.
For the specific variant considered here, the analysis of \cite{BCB12} gives:

\begin{theorem}[\ac{UCB} algorithm for stochastic bandits]
\label{thm:UCB}
The \ac{UCB} algorithm with parameter $\alpha>2$ enjoys the worst-case guarantee
\begin{equation}
\label{eq:reg-UCB}
\bar\reg_{\horizon}
	\leq \sum_{\pure\neq\opt\pure} \parens*{\frac{2\alpha}{\gap_{\pure}}  \log\horizon + \frac{\alpha}{\alpha-2}}
	= \bigoh(\log\horizon).
\end{equation}
\end{theorem}

Importantly, \hilite{the regret guarantee of \ac{UCB} is optimal in $\horizon$:}
the information-theoretic analysis of Lai and Robbins shows that no causal policy played against a bandit with Bernoulli reward distributions can achieve regret lower than $\Omega(\log\horizon)$ \cite{LR85}.
Eliminating the gap between the multiplicative constants that appear in \eqref{eq:reg-UCB} and the Lai\textendash Robbins lower bound has given rise to a significant body of literature:
we only mention here the recent \acs{KL}\textendash\ac{UCB} variant of \cite{GC11} which essentially attains the Lai\textendash Robbins bound and strictly dominates \eqref{eq:UCB} for any class of bounded reward distributions.
Further refinements under different assumptions are also possible, and we refer the reader to \cite{BCB12} for an introduction to this rich literature.
Other than that however, in an application-agnostic setting, \ac{UCB} provides a fairly complete solution to the stochastic \ac{MAB} problem.

\subsection{Adversarial Bandits}
\label{sec:bandits-adv}

Stochastic bandits have found a wide range of applications in problems with a clear statistical character such as filtering \cite{GH13}, sequence prediction \cite{CBL06}, etc.
However, when the problem at hand is not purely statistical in nature (or when its statistics are influenced by exogenous, contextual factors), it is no longer possible to identify an ``optimal'' arm.
In the context of \acl{SP}, this is particularly important in communication problems that involve (slow) fading or jamming-proof coding \cite{MB16},
generative neural networks for image processing \cite{GPM+14},
game-theoretic applications \cite{MZ16u},
etc.

Tracing its roots to the ``rigged casino'' problem of \cite{ACBFS02}, this paradigm is known as \emph{adversarial} because the agent is called to learn against \emph{any} possible sequence of rewards, including those designed by a mechanism that actively tries to minimize the agent's cumulative payoff over time \textendash\ an \emph{adversary}.
More precisely, at each step in an adversarial bandit problem, the rewards
$\pay_{\run}(\pure)$ of each arm $\pure\in\pures$ are determined by the adversary
simultaneously with the agent's action $\pure_{\run}$, possibly with full knowledge of the decision process employed by the agent at step $\run$.
Thus, optimizing in hindsight over the horizon of play, the agent's realized regret is now defined as
\begin{equation}
\label{eq:reg-realized}
\reg_{\horizon}
	= \max_{\pure\in\pures} \sum_{\run=1}^{\horizon} \bracks{\pay_{\run}(\pure) - \pay_{\run}(\pure_{\run})},
\end{equation}
\ie as the cumulative payoff difference between the agent's chosen arm and the best possible arm over the given horizon of play.

\subsubsection{Adversarial vs. stochastic bandits}

In an adversarial setting, the agent's objective is to incur sublinear regret $\reg_{\horizon} = o(\horizon)$ against \emph{any} sequence of rewards chosen by the adversary.
However, in contrast to the stochastic case, this definition is not inherently probabilistic and does not include any (stochastic) averaging.
In particular, if
\begin{inparaenum}
[\itshape i\hspace*{1pt}\upshape)]
\item
the agent's choice of arm is made by a deterministic algorithm that takes as input the history of play up to the current stage;
and
\item
the adversary knows this algorithm in advance and uses it to design the worst possible sequence of outcomes,
\end{inparaenum}
the sequence of play and the generated rewards are all deterministic.

This simple example shows that the stochastic and adversarial bandit paradigms are not immediately comparable.
More importantly, it also illustrates \hilite{Cover's impossibility result} \cite{Cov65}:
\emph{a deterministic algorithm cannot hope to achieve sublinear regret against an adversarial bandit.}
Indeed, if the optimizer uses a deterministic algorithm which is known to the adversary, the adversary could set the reward $\pay_{\iRun}(\pure_{\iRun})$ of the optimizer's chosen arm $\pure_{\iRun}$ to $0$, and the rewards of all other arms $\pure\neq\pure_{\iRun}$ to $1$.
In so doing, after $\horizon$ stages, the agent will have gained a cumulative payoff of $0$;
however, any arm $\pure\in\pures$ which has been chosen no more than $\horizon/2$ times would have generated a cumulative payoff of at least $\horizon/2$.%
\footnote{That such an arm exists follows from the pigeonhole principle:
it is not possible for all arms to be chosen more than $\horizon/2$ times.}
The incurred regret is therefore $\reg_{\horizon} \geq \horizon/2$, showing that the agent cannot achieve sublinear regret in this setting.

Cover's result shatters all hopes of minimizing regret with a deterministic algorithm in an adversarial setting.
However, if the choice of the arms is \emph{randomized}, different results may hold against an adversary.
By this token, a meaningful relaxation of the problem is to allow for \emph{randomized arm selection:}
in this case, the agent is choosing an action $\pure_{\run}$ at stage $\run$ based on a probability distribution (or \emph{mixed strategy}) $\act_{\run} = (\act_{\pure,\run})_{\pure\in\pures}$
which assigns probability $\act_{\pure,\run}$ to arm $\pure$.
The near-omniscient adversary might be fully aware of the agent's mixed strategy at stage $\run$, but the actual results of these randomized draws are not known in advance \textendash\ otherwise, Cover's impossibility result would still apply.
The agent's mean regret is then defined as
\begin{equation}
\label{eq:reg-mean-adv}
\bar\reg_{\horizon}
	= \max_{\pure\in\pures} \sum_{\run=1}^{\horizon} \exof*{\pay_{\run}(\pure) - \pay_{\run}(\pure_{\run})}
\end{equation}
where
the expectation $\exof{\cdot}$ is taken over the optimizer's mixed strategy $\act_{\run}\in\simplex(\pures)$, $\run=1,\dotsc,\horizon$,
and any randomization employed by the adversary (assumed here to be independent of the agent's randomized outcomes).

A first observation is that if the adversary chooses the rewards of each arm by drawing an \ac{iid} random variable from some fixed distribution, \cref{eq:reg-mean,eq:reg-mean-adv} for the stochastic and adversarial setting coincide:
in view of this,
\hilite{any regret guarantee for adversarial bandits translates to a regret guarantee for stochastic bandits.}
However, given that \ac{UCB} is inherently deterministic (in that the arm chosen at each step is dictated by a deterministic criterion), this transfer of results does not go the other way:
\hilite{an algorithm attaining no regret in the stochastic framework may still incur significant regret against an adversary.}

\subsubsection{No regret in adversarial environments}

Going back to the original ``rigged casino'' setting of \cite{ACBFS02}, a key idea in balancing exploration versus exploitation in the adversarial setting is to keep a cumulative score of the performance of each arm and then employ an arm with probability that is exponentially proportional to this score.
Assuming for the moment that the optimizer has \emph{full information} \textendash\ \ie they can observe the bandit's entire payoff vector $\payv_{\run} = (\pay_{\run}(\pure))_{\pure\in\pures}$ after choosing an arm at stage $\run$ \textendash\ this leads to the so-called ``Hedge'' or \acdef{EW} algorithm:%
\footnote{Depending on the amount of exploration and the ``weights'' involved in the \ac{EW} algorithm, there are several variants thereof such as the \ac{MW} algorithm, the \ac{EXP3}, and others.}
\begin{equation}
\label{eq:EW}
\begin{aligned}
\score_{\run+1}
	&= \score_{\run} + \step \payv_{\run},
	\\
\act_{\run+1}
	&= \logit(\score_{\run+1}),
\end{aligned}
\end{equation}
where the \emph{logit choice map} $\logit\from\R^{\pures}\to\simplex(\pures)$ is given by
\begin{equation}
\label{eq:logit}
\logit(y)
	= \frac{(\exp(y_{\pure}))_{\pure\in\pures}}{\sum_{\pure\in\pures} y_{\pure}}.
\end{equation}


\begin{algorithm}[tbp]
\caption{Hedge/\acf{EW}}
\label{alg:EW}
\input{Algorithms/EW}
\end{algorithm}


The \ac{EW} algorithm has close links with the \emph{replicator dynamics} in evolutionary biology and reinforcement learning in games \cite{MS16}.
Heuristically, the algorithm reinforces the probability of choosing an arm that has performed well in the past while maintaining the possibility of exploring other arms (which may perform better in the future). 
This exploration-exploitation tradeoff is tuned here by the parameter $\step>0$:
For very small values of $\step$, the differences between scores are negligible, so all arms are drawn with roughly uniform probability.
On the other hand, for large values of $\step$, payoff differences are compounded by the \ac{EW} algorithm, so the agent tends to select the highest-scoring arm with overwhelming probability.
Making this tradeoff more precise, Auer et al. \cite{ACBFS02} obtained the regret guarantee
\begin{equation}
\label{eq:reg-EW}
\bar\reg_{\horizon}
	\leq \frac{\log\nPures}{\step} + \frac{1}{2} \step\horizon.
\end{equation}

This bound illustrates the delicate role played by $\step$ in the dilemma between exploration and exploitation:
when $\step$ is small, the ``exploration regret'' $\step\horizon$ is minimized, but the ``exploitation regret'' $\log\nPures/\step$ is maximized;
conversely, for large values of $\step$, the exploration regret $\step\horizon$ is maximized, while the exploitation regret $\log\nPures/\step$ is minimized.
Thus, optimizing the choice of $\step$ as a function of $\horizon$, we have

\begin{theorem}[\ac{EW} algorithm with full information]
\label{thm:EW}
The \ac{EW} algorithm with learning rate $\step = \sqrt{2\log\nPures / \horizon}$ achieves the regret bound
\begin{equation}
\label{eq:reg-EW2}
\bar\reg_{\horizon}
	\leq \sqrt{2\horizon\log\nPures}
	= \bigoh\parens*{\sqrt{\horizon\log\nPures}}.
\end{equation}
\end{theorem}

Given the adversary's near-omniscience, the fact that a minimal amount of randomization allows the agent to ``surprise'' the adversary often enough to attain a no-regret state is quite remarkable.%
\footnote{In \cref{sec:algorithms}, we will see that one explanation of this is that randomization ``convexifies'' the agent's rewards.}
Under this light, the added challenge of gambling against a rigged casino (as opposed to an ordinary stochastic bandit) is only reflected in the performance drop from $\bigoh(\log\horizon)$ in the stochastic setting to $\bigoh(\horizon^{1/2})$ in the adversarial case.
That said, this gap is effectively insurmountable:
Cesa-Bianchi et al. \cite{CBFH+97} showed that, for any causal algorithm with full information, there exists an adversarial sequence of payoffs such that
\begin{equation}
\label{eq:reg-lower-adv-full}
\bar\reg_{\horizon}
	= \Omega\parens*{\sqrt{\horizon\log\nPures}}.
\end{equation}
In other words,
\hilite{there is a gap of the order of $\Omega(\sqrt{\horizon}/\log\horizon)$ between stochastic and adversarial bandits, even with full information for the latter.}


\begin{figure*}[t]
\footnotesize
\subfigure{\includegraphics[width=.48\textwidth]{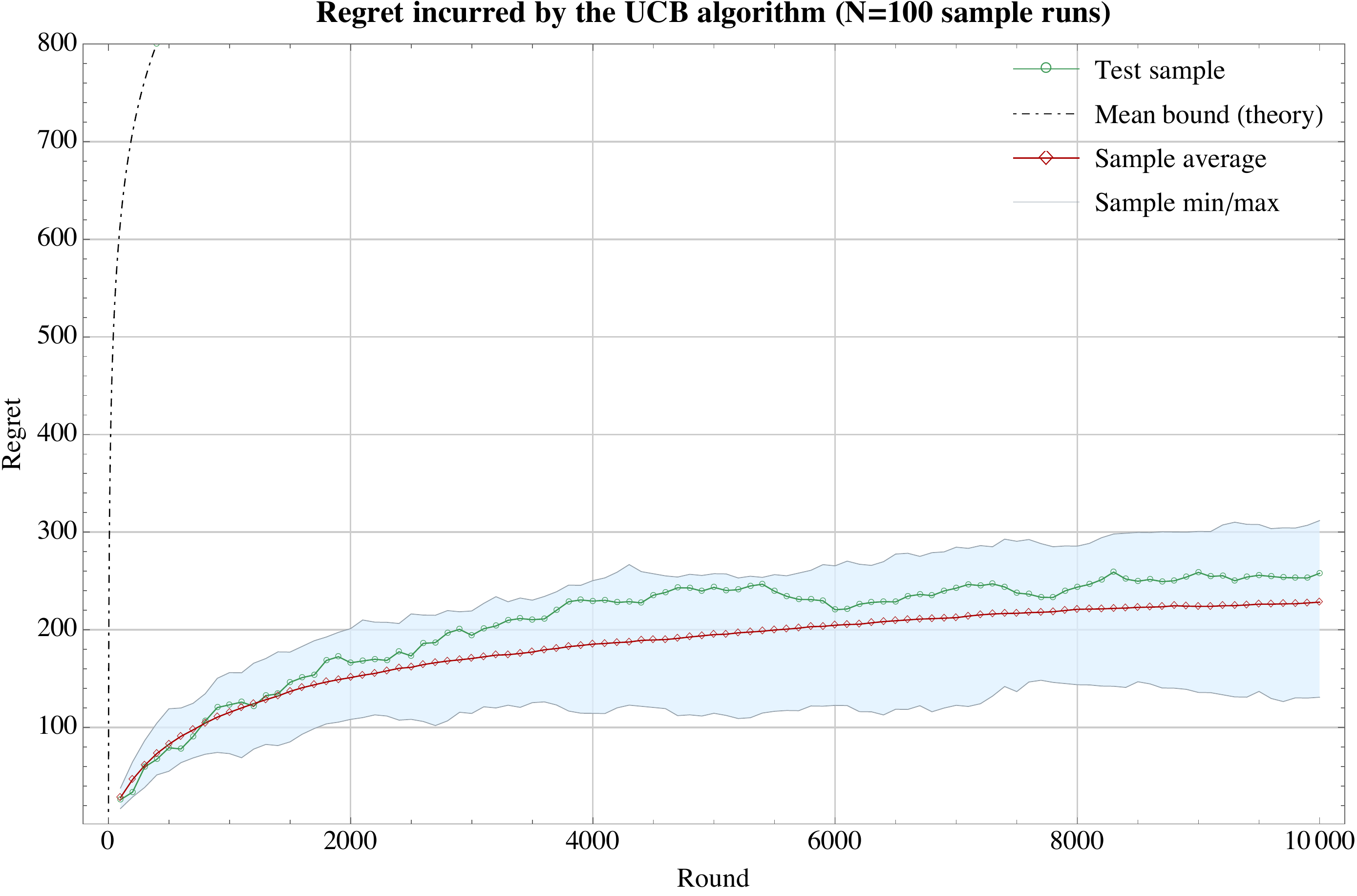}}
\hfill
\subfigure{\includegraphics[width=.48\textwidth]{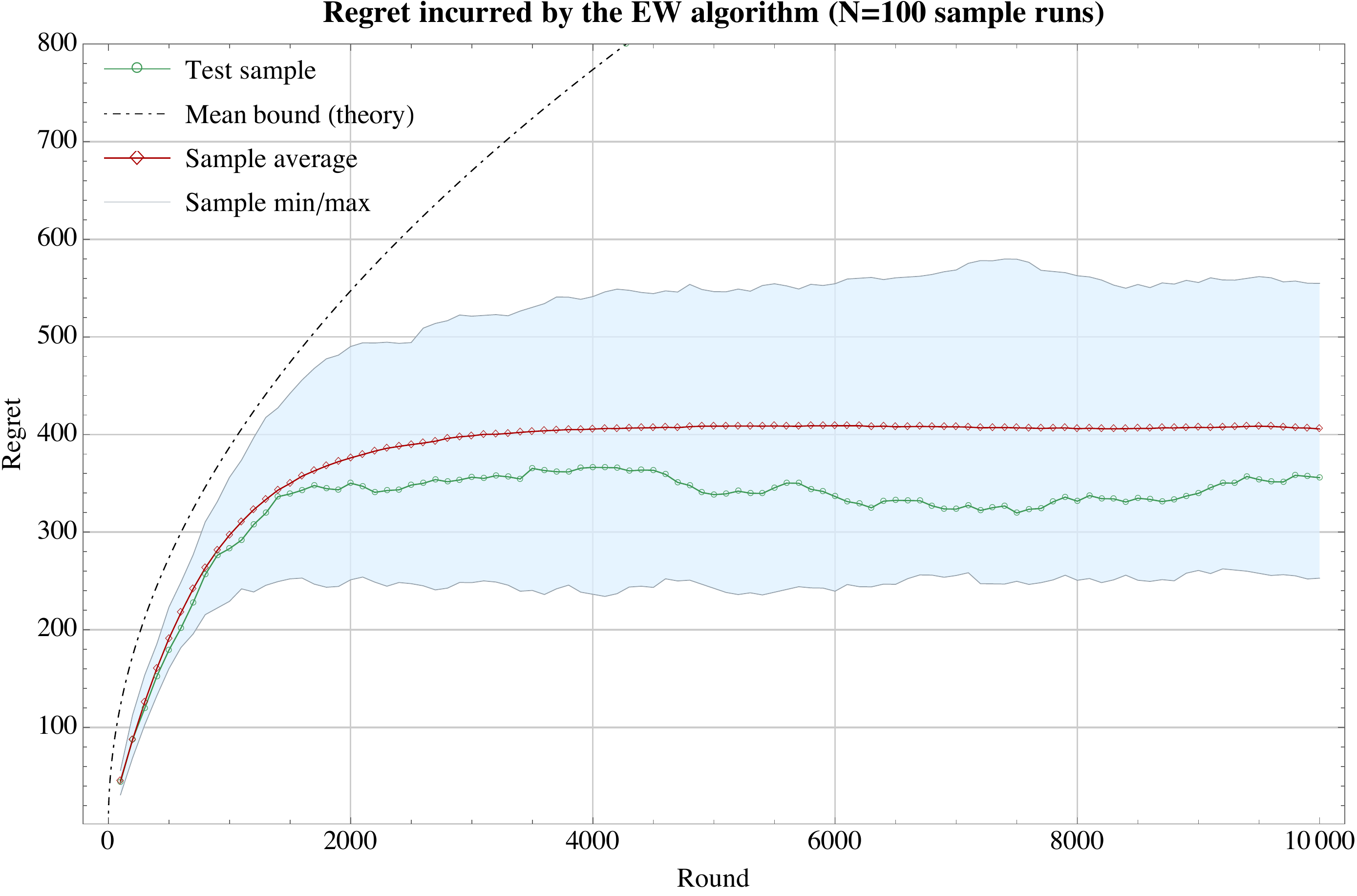}}
\caption{Performance of the \ac{UCB} and \ac{EW} algorithms in a channel selection problem with $\nChannels=16$ channels and random occupancy statistics.
The user's regret (measuring the throughput gap against the best channel) quickly stabilizes, giving an error probability of as little as $1\%$ relative to the best channel.}
\label{fig:UCB-EXP}
\end{figure*}


\subsubsection{Full vs. partial information}

From an information-theoretic point of view, a key difference between the \ac{UCB} and \ac{EW} algorithms is that the former does not require any further information except for the agent's realized payoff $\hat\pay_{\run} = \payv_{\pure_{\run},\run}$ (the payoff obtained by choosing the arm $\pure_{\run}$) as opposed to relying also on the knowledge of the unchosen arms.
Thus, since the $\bigoh(\sqrt{T})$ regret guarantee of the \ac{EW} algorithm relies on full information, a key question remains:
\hilite{is it possible for the agent to attain a no-regret state in an adversarial environment with partial information?}

To address this issue, the first step is to construct an estimator for the bandit's payoff vector based on observations of the agent's realized payoffs.
This is often achieved by the so-called \hilite{importance sampling} technique:
if the payoff vector at stage $\run$ is $\payv_{\run}$ and the agent gains $\hat\pay_{\run} = \payv_{\pure_{\run},\run}$ after choosing arm $\pure_{\run}$, the otherwise unobserved payoff vector can be estimated by setting 
\begin{equation}
\label{eq:importance}
\hat\payv_{\pure,\run}
	= \begin{cases}
		\hat\pay_{\run} / \act_{\pure,\run}
		&\quad
		\text{if $\pure=\pure_{\run}$},
		\\
		0
		&\quad
		\text{otherwise},
	\end{cases}
\end{equation}
where $\act_{\pure,\run} = \probof{\pure_{\run} = \pure}$ is the probability with which arm $\pure$ is drawn based on the agent's mixed strategy $\act_{\run}$ at stage $\run$.
The key property of the resulting estimator $\hat\payv_{\run}$ is that it is \emph{unbiased}, \ie
\begin{equation}
\label{eq:unbiased}
\exof{\hat\payv_{\run}}
	= \payv_{\run}.
\end{equation}
As such, even if the optimizer can only observe their realized payoffs, it is possible to run the \ac{EW} algorithm with $\hat\payv_{\run}$ in place of $\payv_{\run}$ (for a pseudocode implementation, see \cref{alg:EW-partial}).


\begin{algorithm}[tbp]
\caption{\acf{EW} with importance sampling}
\label{alg:EW-partial}
\input{Algorithms/EW-partial}
\end{algorithm}


Doing so, a finer regret analysis reveals that the \ac{EW} algorithm with partial information enjoys the regret bound \cite{SS11}
\begin{equation}
\label{eq:reg-EW-partial}
\bar\reg_{\horizon}
	\leq \frac{\log\nPures}{\step} + \step\nPures\horizon.
\end{equation}
Comparing this guarantee to the full information bound \eqref{eq:reg-EW}, we see that the price of learning with partial information is an effective rescaling of the horizon $\horizon$ to $\nPures\horizon$.
In a certain sense, this is to be expected as the agent now has to play $\bigoh(\nPures)$ more times in order to get a better handle on the bandit's payoff vector.
More concretely, if the parameter $\step$ is optimized in terms of $\nPures$ and $\horizon$, we have \cite{SS11}:

\begin{theorem}[\ac{EW} algorithm with partial information]
\label{thm:EW-partial}
The \ac{EW} algorithm with partial information and learning rate $\step = \sqrt{\log\nPures / (\nPures\horizon)}$ enjoys the regret bound
\begin{equation}
\label{eq:reg-EW2-partial}
\bar\reg_{\horizon}
	\leq 2\sqrt{\nPures\horizon\log\nPures}
	= \bigoh\parens*{\sqrt{\nPures \horizon \log\nPures}}.
\end{equation}
\end{theorem}

The above bound shows that the passage from full to partial information in an adversarial environment worsens the incurred regret of the \ac{EW} algorithm by a factor of $\sqrt{\nPures}$.
This gap can be tightened, but only slightly so:
an information-theoretic analysis of the adversarial bandit problem \cite{ACBFS02} reveals that the mean regret of any causal algorithm with partial information is bounded from below as
\begin{equation}
\label{eq:reg-lower-adv-partial}
\bar\reg_{\horizon}
	= \Omega\parens*{\sqrt{\nPures\horizon}}.
\end{equation}
Thus, comparing \cref{eq:reg-lower-adv-full,eq:reg-lower-adv-partial}, we see that it is not possible to bridge the gap between full and partial information when playing against an adversary.

In recent work, Audibert and Bubeck \cite{AB10} shaved off the logarithmic gap between the upper and lower bounds \eqref{eq:reg-EW2-partial} and \eqref{eq:reg-lower-adv-partial} by tweaking the exponential map in \eqref{eq:EW} and employing a so-called \acdef{INF}.
As in the stochastic case, further refinements are possible depending on what assumptions are made for the adversary \textendash\ see \eg \cite{CBL06} for some variants.
However, except for such refinements, the \ac{EW} algorithm essentially provides a complete solution to adversarial bandit problems;
the various bounds are summarized in \cref{tab:reg-bandits}.


\begin{table}[tbp]
\centering
\renewcommand{\arraystretch}{1.2}
\input{Tables/Bounds-Bandits}
\caption{Lowest achievable (minimax) regret bounds in \aclp{MAB} and algorithms that attain them (up to logarithmic factors in $\nPures$).}
\label{tab:reg-bandits}
\end{table}


%% file: Algorithms/UCB.tex

\begin{algorithmic}[1]
\REQUIRE
	tuning parameter $\alpha>2$,
	initial reward sample $\hat\mean_{\pure}$ from each arm $\pure\in\pures$
\STATE
	set $\nTimes_{\pure} = 1$ for all $\pure\in\pures$
	\COMMENT{initialization}%
\FOR{$\run=1$ \TO $\horizon$}
	\STATE
		draw $\pure_{\run} \in \argmax_{\pure\in\pures} \braces*{\hat\mean_{\pure} + \sqrt{\alpha\log\run/(2\nTimes_{\pure})}}$
		\COMMENT{arm selection}%
	\STATE
		gain $\hat\pay_{\run} = \payv_{\pure_{\run},\run}$
		\COMMENT{receive reward}%
	\STATE
		$\nTimes_{\pure_{\run}} \leftarrow \nTimes_{\pure_{\run}} + 1$
		\COMMENT{sample size update}%
	\STATE
		$\hat\mean_{\pure_{\run}} \leftarrow \parens*{1 - \frac{1}{\nTimes_{\pure_{\run}}}} \, \hat\mean_{\pure_{\run}} + \frac{1}{\nTimes_{\pure_{\run}}} \hat\pay_{\run}$
		\COMMENT{empirical mean update}%
\ENDFOR
\end{algorithmic}

%% file: Algorithms/EW.tex

\begin{algorithmic}[1]
\REQUIRE
	parameter $\step>0$
\STATE
	set $\score_{\pure} \leftarrow 0$ for each arm $\pure\in\pures$
	\COMMENT{initialization}%
\FOR{$\run=1$ \TO $\horizon$}
	\STATE
		play $\act \leftarrow \logit(\score)$
		\COMMENT{logit update as in \eqref{eq:logit}}%
	\STATE
		draw $\pure_{\run}$ according to $\act$
		\COMMENT{arm selection}%
	\STATE
		get payoff vector $\payv_{\run}$ and receive $\hat\pay_{\run} = \payv_{\run,\pure_{\run}}$
		\COMMENT{payoffs revealed}%
	\STATE
		set $\score \leftarrow \score + \step\payv_{\run}$
		\COMMENT{score update}%
\ENDFOR
\end{algorithmic}

%% file: Algorithms/EW-partial.tex

\begin{algorithmic}[1]
\renewcommand{\algorithmicrequire}{\textbf{Replace}}
\REQUIRE
	the following lines in \cref{alg:EW}
\makeatletter
\setcounter{ALC@line}{4}
\makeatother
	\STATE
		receive $\hat\pay_{\run} = \payv_{\run,\pure_{\run}}$ and set $\hat\payv_{\pure} \leftarrow \hat\pay_{\run} \cdot \one(\pure = \pure_{\run}) / x_{\pure}$ for all $\pure\in\pures$
		\COMMENT{payoff vector estimate}%
	\STATE
		set $\score \leftarrow \score + \step\hat\payv$
		\COMMENT{score update}%
\end{algorithmic}

%% file: Tables/Bounds-Bandits.tex

\small
\begin{tabular}{rllr}
\;
	&\scshape
	Minimax Regret
	&\scshape
	Achieved by
	\\
\hline
\scshape
Stochastic Bandit
	&$\Omega(\log\horizon)$
	&\ac{UCB}
	\\
\hline
\scshape
Adversarial (Full Info)
	&$\Omega\parens{\sqrt{\horizon}}$
	&\ac{EW}
	\\
\hline
\scshape
Adversarial (Partial Info)
	&$\Omega\parens{\sqrt{\nPures\horizon}}$
	&\ac{EW} $+$ importance sampling
	\\
\hline
\end{tabular}
\vspace{2ex}

%% file: Setting.tex

Albeit powerful, the \acl{MAB} framework relies on the important assumption that the agent's actions are drawn from a \emph{finite} set,
an assumption which is not realistic in many cases of practical interest in signal processing.
This calls for a new way of optimizing the agent's choices over time;
this is offered by the framework of \emph{online optimization}, which we motivate and discuss below.

\subsection{Elements of Convex Analysis}
\label{sec:convex}

Much of the discussion to follow will be focused on problems with a convex structure.
We thus begin by recalling some basic elements of convex analysis \textendash\ for a comprehensive primer, \cf \cite{BV04};
see also the recent tutorials \cite{MB10} and \cite{SPFP10} for a signal processing viewpoint.

First, a subset $\feas$ of a $\vdim$-dimensional real space $\R^{\vdim}$ is called \emph{convex} if, for all $\act,\actalt\in\feas$, the line segment $[\act,\actalt] \equiv \setdef{(1-\theta)\act + \theta\actalt}{0\leq\theta\leq1}$ also lies in $\feas$.
If $\feas$ is convex, a function $\obj\from\feas\to\R$ is called convex if
\begin{equation}
\label{eq:cvx}
\obj((1-\theta)\act + \theta\actalt)
	\leq (1-\theta) \obj(\act) + \theta \obj(\actalt)
	\quad
	\text{for all $\act,\actalt\in\feas$ and all $\theta\in[0,1]$},
\end{equation}
\ie when the line segment between any two points on the graph of $\obj$ lies above the graph itself.
When \eqref{eq:cvx} holds as a strict inequality for all $\actalt\neq\act$, the function $\obj$ is called \emph{strictly convex}.
As a refinement of strict convexity, $\obj$ is called \emph{strongly convex} when the gap in the inequality \eqref{eq:cvx} is bounded by a quadratic function, \ie there exists some $\strong>0$ such that
\begin{equation}
\label{eq:cvx-strong}
\obj((1-\theta)\act + \theta\actalt)
	\leq (1-\theta) \obj(\act) + \theta \obj(\actalt)
	- \frac{\strong}{2} \theta(1-\theta) \norm{\actalt-\act}^{2},
\end{equation}
or, equivalently, if the function $\obj(x) - \tfrac{1}{2}\strong \norm{\act}^{2}$ is convex.
If $\obj$ is sufficiently differentiable, this curvature requirement is captured by asking that the Hessian of $\obj$ be bounded from below as $\Hess(\obj) \mgeq \strong\,\bI$.

Taking $\feas=\R^{\vdim}$, examples of convex functions include
\begin{inparaenum}
[\itshape i\upshape)]
\item
affine functions of the form $\obj(\act) = \product{a}{\act} + b$ for $a,b\in\R^{\vdim}$;
\item
quadratics of the form $\obj(\act) = \act^{\top} \bM \act$ for some positive-definite $\bM \in \R^{\vdim\times\vdim}$;
\item
the log-sum-exp function $\obj(\act) = \log\sum_{j=1}^{\vdim} \exp(\act_{j})$ which is widely used in dictionary learning and image processing;
etc.
\end{inparaenum}
Of these examples,
affine functions are convex but not strictly convex,
quadratics are strongly convex when $\bM\mg0$,
and the log-sum-exp function is strictly convex but not strongly convex.

\subsection{Online Optimization}
\label{sec:online}

With these preliminaries at hand, we are ready to describe our core online optimization framework:
At each stage $\run=1,2,\dotsc$, the optimizing agent selects an action $\act_{\run}$ from some compact convex set $\feas\subseteq\R^{\vdim}$ and incurs a loss of $\loss_{\run}(\act_{\run})$ based on some (a priori unknown) loss function $\loss_{\run}\from\feas\to\R$
(as in the bandit case, this loss function could be determined stochastically, adversarially, or otherwise).
Subsequently, the agent selects a new action $\act_{\run+1}$ for the next stage and the process repeats as shown below:


\begin{process}
\label{alg:ODP}
\small
\caption{\bfseries Generic online decision process}
\input{Algorithms/ODP}
\end{process}


In the above description of online decision processes, no assumptions are made about the structure of the loss functions $\loss_{\run}$.
Typically, depending on their properties, an appropriate adjective is affixed to the term ``online optimization'';
in particular, we have the following fundamental classes of online optimization problems:
\begin{itemize}
\item
\emph{Online convex optimization:}
$\loss_{\run}\from\feas\to\R$ is convex for all $\run$.
\item
\emph{Online strongly convex optimization:}
$\loss_{\run}\from\feas\to\R$ is strongly convex for all $\run$.
\item
\emph{Online linear optimization:}
$\loss_{\run}$ is of the form $\loss_{\run}(\act) = -\product{\payv_{\run}}{\act}$ for some vector $\payv_{\run}\in\R^{\vdim}$ and all $\run$.
\end{itemize}

Both linear and strongly convex problems form subclasses of the convex class, but they are obviously pairwise disjoint.
Other classes of problems have also been studied in the literature \cite{CBL06}, but the above are the most common ones by far.
For illustration purposes, we discuss some important examples below.

\subsection{Examples of Online Optimization Problems}
\label{sec:examples}

\subsubsection{Multi-armed bandits revisited}

Going back to the discussion of the previous section, let $\feas = \simplex(\pures) = \setdef{\act\in\R_{+}^{\pures}}{\sum_{\pure\in\pures} \act_{\pure} = 1}$ denote the simplex of probability distributions over the set of arms of a \acl{MAB}.
If the agent uses mixed strategy $\act_{\run}\in\feas$ at stage $\run$ and the bandit's reward vector is $\payv_{\run}$, the agent's mean reward will be $\product{\payv_{\run}}{\act_{\run}}$.
Hence, in a loss-based formulation, the agent will be incurring a mean loss of
\begin{equation}
\label{eq:loss-MAB}
\loss_{\run}(\act_{\run})
	= - \product{\payv_{\run}}{\act_{\run}}.
\end{equation}
Thus, going back to \eqref{eq:reg-mean-adv} the agent's (mean) regret can be reformulated as
\begin{equation}
\bar\reg_{\horizon}
	= \max_{\pure\in\pures} \sum_{\run=1}^{\horizon} \exof*{\pay_{\run}(\pure) - \pay_{\run}(\pure_{\run})}
	= \max_{\act\in\simplex(\pures)} \sum_{\run=1}^{\horizon} \product{\payv_{\run}}{(\act - \act_{\run})}
	= \max_{\act\in\feas} \sum_{\run=1}^{\horizon} \bracks{\loss_{\run}(\act_{\run}) - \loss_{\run}(\act)}.
\end{equation}
This gives us an alternative way of interpreting the incurred regret in a \acl{MAB} problem:
it is simply the difference between the aggregate loss incurred by the agent's chosen policy and that of the best mixed strategy in hindsight.
Since losses are linear, maximizing over the entire simplex (mixed strategies) or its vertices (pure strategies) does not change anything, so the two definitions coincide.
As such, all applications of \ac{MAB} problems discussed in Sec. \ref{sec:bandits} fall under the umbrella of online optimization.

\subsubsection{Multi-antenna communications}
\begin{figure}[tbp]
\centering
\input{Figures/MU-MIMO-network.tex}
\caption{A \ac{MU}-\ac{MIMO} wireless network:
solid blue arrows indicate the intended receiver, dashed red arrows indicate interference crosstalk.}
\label{fig:mu-mimo}
\end{figure}
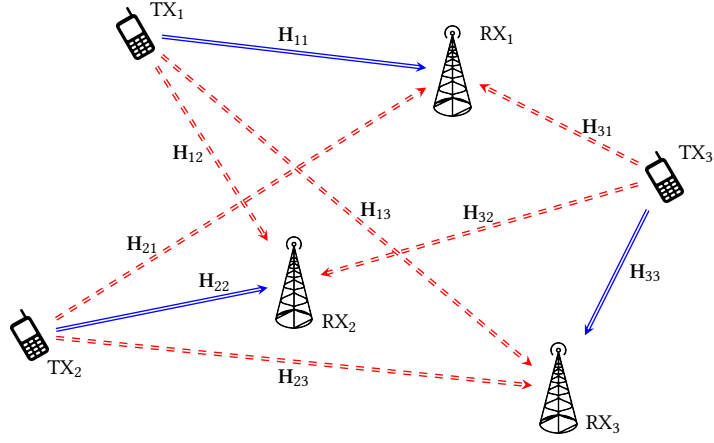

The next example concerns the problem of signal covariance optimization in \ac{MIMO} wireless networks \cite{SPFP10,MM16,MBNS17}.
To state it, consider the dynamic multi-user \ac{MIMO} wireless network of \cref{fig:mu-mimo} where a set of autonomous wireless devices equipped with multiple antennas seek to maximize their individual data rates.
Focusing on a single transmitter for convenience, its Shannon achievable rate is given by
\begin{equation}
\label{eq:Shannon}
\rate(\bQ;\tilde\bH)
	= \log\det(\bI + \tilde\bH \bQ \tilde\bH^{\dag}),
\end{equation}
where $\bQ$ denotes the signal covariance matrix of the focal transmitter and $\tilde\bH$ its effective gain matrix \cite{MM16}.%
\footnote{For instance, the effective channel matrix of the link between transmitter $\mathrm{TX}_1$ and receiver $\mathrm{RX}_1$ (user $1$) in \cref{fig:mu-mimo} is $\tilde \bH_{1} = \bW_1^{-1/2} \bH_{11}$ where $\bW_1 = \mathbf{\Sigma_1} + \bH_{21} \bX_{2}  \bH_{21}^{\dag}+\bH_{31} \bX_{3}  \bH_{31}^{\dag}$ is the multi-user interference plus noise covariance matrix, $\mathbf{\Sigma_1}$ is the received noise covariance matrix, $\bX_{2}$, $\bX_{3}$ are the transmit covariance matrices of the interfering users, and $\bH_{ij}$ are the channel matrices between $\mathrm{TX}_i$ and $\mathrm{RX}_j$.}
By standard considerations, $\rate$ is concave in $\bQ$, so its maximization is relatively easy if $\tilde\bH$ is known in advance.
However, since $\tilde\bH$ encompasses the effects of the wireless medium (noise, pathloss, device mobility) and depends on the transmit characteristics of all interfering users (which may go on- and off-line in an ad-hoc manner), it cannot be known in advance in a time-varying environment. 
Thus, letting $\pmax$ denote the user's maximum transmit power and writing $\feas = \setdef{\bQ\mgeq0}{\tr\bQ\leq\pmax}$ for the set of possible covariance matrices, rate maximization in a dynamic \ac{MIMO} network can be formulated as an online convex optimization problem where the user seeks to minimize the incurred loss $\loss_{\run}(\bQ_{\run}) = - \rate(\bQ_{\run};\tilde\bH_{\run})$ against any possible sequence of dynamically determined $\tilde\bH_{\run}$.
Energy efficiency maximization \cite{MB16} and cognitive medium access \cite{SP10,MB14} are other important applications in \ac{MIMO} systems that can be formulated as online optimization problems.

\subsubsection{Multimedia indexing}
\label{sec:online-metric}

The last example comes from supervised metric learning for image similarity search and classification \cite{Negrel-2013, Bellet-2013}.
The aim here is to learn a positive-definite matrix $\bX$ shaping the \emph{Mahalanobis distance} $d_{X}(\bp, \bq)$ that best captures the similarity of two images $\bp$ and $\bq$, based upon existing annotations or examples (supervised learning);
concretely, we have:
\begin{equation}
\label{eq:Mahalanobis}
d_{X}(\bp, \bq) = (\bp-\bq)^{\top} \bX (\bp - \bq).
\end{equation}

An example is a triplet $(\bp,\bq, y)$ where $\bp$, $\bq$ are two images and $y$ represents their similarity score (so $y=+1$ if the images are similar and $y=-1$ otherwise).
Intuitively, the Malahanobis distance performs a linear transformation of the data and computes the distance $d_{X}(\bp, \bq) = \norm{\bX^{1/2} \bp - \bX^{1/2} \bq}^{2}$ in the transformed space.
Then, the objective is to learn the best transformation that brings closer the similar images and separates the dissimilar ones.

Following \cite{Kunapuli2012}, this can be formalized by asking that 
\begin{itemize}
\item
For any example $(\bp,\bq, +1)$ of similar images $\bp$ and $\bq$, the distance separating them should be relatively small, i.e., $d_{X}(\bp, \bq) \leq \eps - 1$.
\item
For any example $(\bp,\bq, -1)$ of dissimilar images images $\bp$ and $\bq$, the distance separating them should be relatively large, i.e., $d_{X}(\bp, \bq) \geq \eps +1$.
\end{itemize}
In the above, $\eps \geq 1$ is a \emph{threshold margin} which has to be learned jointly with $\bX$.
The above requirements amount to satisfying the compact similarity constraint
\begin{equation}
\label{eq:sim_cond}
y(\eps - d_{X}(\bp, \bq) ) \geq 1, 
\end{equation}
which should be met for all examples $(\bp,\bq,y)$.


\begin{figure}[tbp]
\centering
\includegraphics[width=0.7\textwidth]{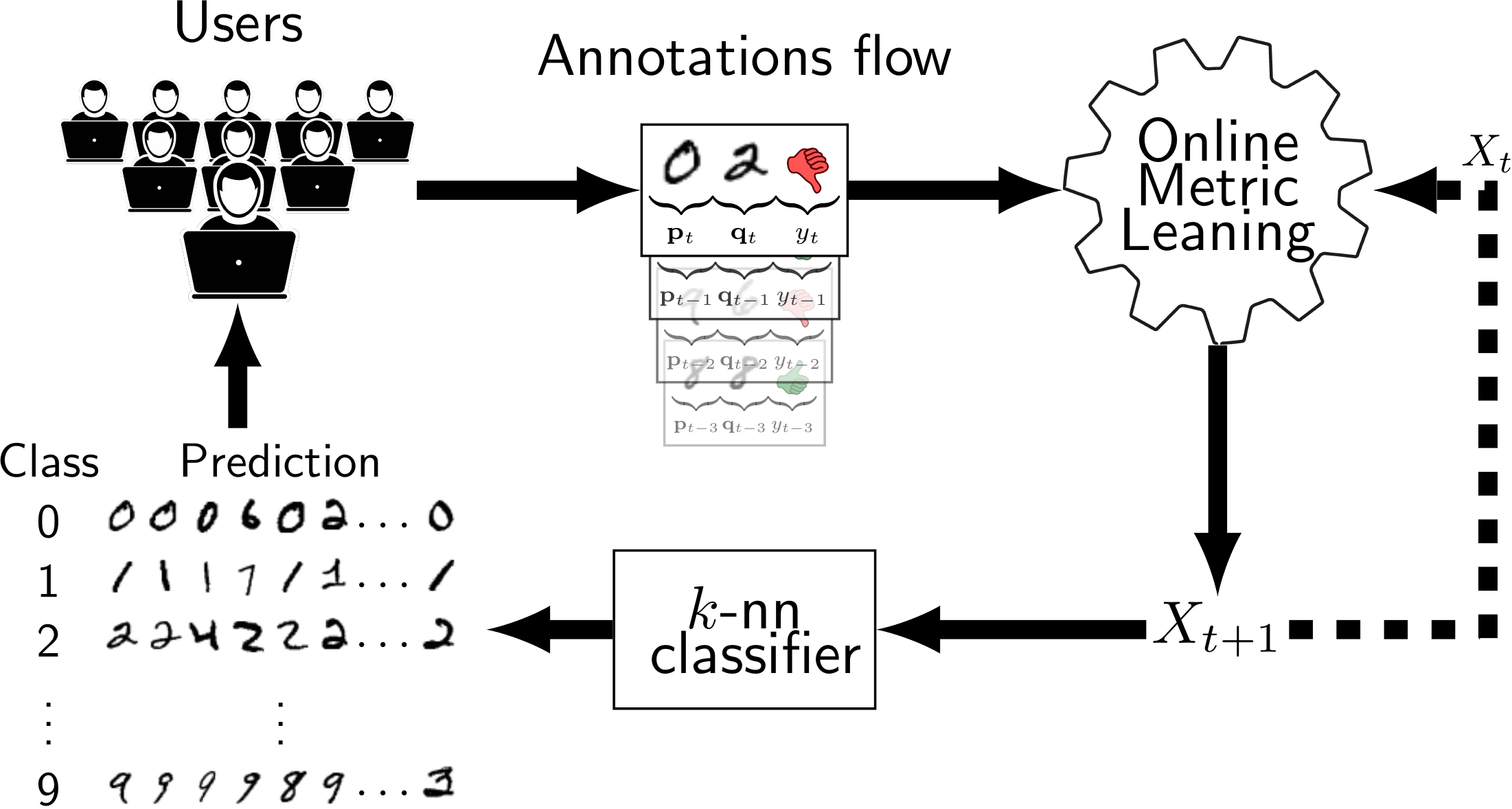}
\caption{Online metric learning for MNIST database digit classification using the $k$-nearest neighbors classifier.}
\label{fig:ml_clasif}
\end{figure}


In online metric learning, the process is based on annotations that are not all available at once but become available on-the-fly (as opposed to batch methods that use offline datasets for training) \cite{Kunapuli2012, Jain2009, Shalev2004}.
This is particularly relevant in Internet applications where the annotations are collected continuously and in a distributed manner:
there, the feedback coming from the users by scoring pairs of images (and effectively providing new examples) is exploited to improve and adapt incrementally the existent metric, as depicted in \cref{fig:ml_clasif}.
Specifically, at each stage $\run$, the process receives a new user annotation $(\bp_{\run}, \bq_{\run}, y_{\run})$ and incurs the hinge loss 
\begin{equation}
\loss_{\run}(\bX_{\run}, \eps_{\run})
	= \max \left\{0, 1- y_{\run}\left( \eps_{t} - d_{X}(\bp_{\run}, \bq_{\run}) \right)  \right\},
\end{equation}
which penalizes the existing measure $\bX_{\run}$ and threshold $\eps_{\run}$ whenever the similarity constraint in \eqref{eq:sim_cond} is violated.
The problem's feasible set is therefore $\feas =\setdef{(\eps, \bX)}{\eps \geq 1, \bX \mgeq 0, \tr\bX\leq c}$, where the trace constraint plays the role of a relaxed rank (or sparsity) constraint enabling the reduction of computational costs, model complexity, and over-fitting effects.
\cref{fig:ml_clasif} depicts a digit classification pipeline investigated in \cite{Kunapuli2012} for the MNIST database (taken from \url{http://yann.lecun.com/exdb/mnist/}) exploiting the online learned metric in the $k$-nearest neighbors classifier.

\subsection{Regret in Online Optimization}
\label{sec:regret}

Recalling the discussion of the previous section, the agent's regret in an online optimization problem is defined as
\begin{equation}
\label{eq:regret}
\reg_{\horizon}
	= \max_{\act\in\feas} \sum_{\run=1}^{\horizon} \bracks{\loss_{\run}(\act_{\run}) - \loss_{\run}(\act)}
	= \sum_{\run=1}^{\horizon} \loss_{\run}(\act_{\run}) - \min_{\act\in\feas} \sum_{\run=1}^{\horizon} \loss_{\run}(\act),
\end{equation}
\ie as the difference between the aggregate loss incurred by the agent after $\horizon$ stages and that of the best action in hindsight.
As in the bandit case, the agent's regret contrasts the performance of the agent's policy $\act_{\run}$ to that of the optimum action
\begin{equation}
\opt\act
	\in\argmin_{\act\in\feas} \sum_{\run=1}^{\horizon} \loss_{\run}(\act)
\end{equation}
which minimizes the total incurred loss over the given horizon of play.

Of course, this optimum action cannot be calculated if the loss functions encountered by the optimizer are not known in advance.
Consequently, the figure of merit in an online optimization problem is to design a strictly causal, online policy that achieves \emph{no regret}, \ie
\begin{equation}
\label{eq:no-reg}
\reg_{\horizon}
	= o(\horizon),
\end{equation}
As in the \acl{MAB} case, online optimization focuses on designing algorithms that attain the best possible regret minimization rate in \eqref{eq:no-reg}, not only in terms of the horizon of play $\horizon$,
but also in terms of the multiplicative constants that come into \eqref{eq:no-reg}, and which depend on the geometry and dimensionality of the optimizer's feasible set $\feas$.
We address this issue in detail in \cref{sec:algorithms};
for now, we close this section with two important remarks relating online optimization to other optimization paradigms:

\paragraph{Links with static optimization}

A static optimization problem in normal form can be stated as
\begin{equation}
\label{eq:opt}
\tag{Opt}
\begin{aligned}
\textrm{minimize}
	&\quad
	\obj(\act)
	\\
\textrm{subject to}
	&\quad
	\act\in\feas.
\end{aligned}
\end{equation}
Viewed as an \emph{online} optimization problem, this would correspond to facing the same loss function $\loss_{\run} = \obj$ at each stage.
In this case, if $\act_{\run}$ is a no-regret policy and $\obj$ is convex, an immediate application of Jensen's inequality shows that the so-called ``ergodic average''
\begin{equation}
\label{eq:ergodic}
\bar\act_{\horizon}
	= \frac{1}{\horizon} \sum_{\run=1}^{\horizon} \act_{\run}
\end{equation}
enjoys the guarantee
\begin{equation}
\obj(\bar\act_{\horizon}) - \min\obj
	\leq \frac{1}{\horizon} \sum_{\run=1}^{\horizon} \bracks{\obj(\act_{\run}) - \min\obj}
	\leq \frac{\reg_{\horizon}}{\horizon}.
\end{equation}

We thus see that $\bar\act_{\horizon}$ converges to the solution set of \eqref{eq:opt} as $\horizon\to\infty$;
moreover, the rate of this convergence is controlled by the regret minimization rate of $\act_{\run}$.
This key property of no-regret policies has been the cornerstone of a vast literature on fast optimization algorithms;
for a recent overview, see \cite{Nes09}.
As an application, in \cref{sec:OMD} (see also \cref{fig:MIMO-rates}), online policies are compared to classical iterative water-filling algorithms in static \ac{MIMO} networks composed of rate-driven multiple interfering users \cite{SPB09-sp, MM16}.
This example highlights the interest of online optimization tools, especially for their theoretical guarantees and convergence properties.
Indeed, providing convergence guarantees for water-filling algorithms may be very challenging.
For instance, the sufficient conditions in \cite{SPB09-sp} roughly require all interfering links to be dominated by the direct links, which can be rather restrictive and not necessary;
in the particular case of multiple access networks (composed of multiple transmitters and one common receiver as opposed to multiple transmit-receive pairs), such conditions can never be met as the interfering links of one transmitter are the direct links of the others.

\paragraph{Links with stochastic optimization}

The normal form of a stochastic optimization problem is
\begin{equation}
\label{eq:opt-stoch}
\tag{Opt-S}
\begin{aligned}
\textrm{minimize}
	&\quad
	\exof{\sobj(\act;\sample)}
	\\
\textrm{subject to}
	&\quad
	\act\in\feas
\end{aligned}
\end{equation}
where $\sobj(\act;\omega)$ is a stochastic objective function that depends on a random variable $\sample$.
The expectation in \eqref{eq:opt-stoch} is usually very difficult to compute (when at all possible) so, when designing algorithms to solve \eqref{eq:opt-stoch}, it is assumed that $\sobj$ is sampled at an \ac{iid} realization $\sample_{\run}$ of $\sample$ at each iteration.
For instance, in offline metric learning for multimedia indexing, all examples are readily available in a training dataset, but they cannot be exploited simultaneously because of their prohibitive number and size \cite{Bellet-2013}.
To counter this, much smaller random population samples are drawn and exploited iteratively, despite the fact that the stochastic average of this sampling procedure cannot be computed.

From an online viewpoint, this corresponds to a sequence of loss functions of the form $\loss_{\run}(\act) = \sobj(\act;\sample_{\run})$.
As we discuss in detail in the next section, when each $\loss_{\run}$ is convex, no regret is typically achieved by
first
bounding the difference $\loss_{\run}(\act_{\run}) - \loss_{\run}(\act)$ from above as $\product{\nabla\loss_{\run}(\act_{\run})}{(\act_{\run} - \act)}$,
and
then
establishing an upper bound of the form
\begin{equation}
\sum_{\run=1}^{\horizon} \product{\nabla\loss_{\run}(\act_{\run})}{(\act_{\run} - \act)}
	\leq V_{\horizon}
\end{equation}
for some sublinear function $V_{\horizon} = o(\horizon)$.
In the stochastic case where $\loss_{\run}(\act) = \sobj(\act;\sample_{\run})$, this translates immediately to the value convergence guarantee
\begin{equation}
\exof{\sobj(\bar\act_{\horizon};\sample)} - \min \exof{\sobj}
	\leq \frac{1}{\horizon}\sum_{\run=1}^{\horizon} \exof{\product{\nabla\loss_{\run}(\act_{\run})}{(\act_{\run} - \act)}}
	\leq \frac{\exof{V_{\horizon}}}{\horizon}.
\end{equation}
Thus, again, under a no-regret policy, the sequence $\bar\act_{\horizon}$ converges to the solution set of \eqref{eq:opt-stoch} and the rate of convergence is controlled by $V_{\horizon}$, the mean regret minimization rate of $\act_{\run}$.

%% file: Algorithms/ODP.tex

\begin{algorithmic}[1]
\REQUIRE
	action set $\feas$, sequence of loss functions $\loss_{\run}\from\feas\to\R$
\FOR{$\run=1$ \TO $\horizon$}
	\STATE
		choose $\act_{\run}\in\feas$
		\COMMENT{action selection}
	\STATE
		incur $\loss_{\run}(\act_{\run})$
		\COMMENT{incur loss}
	\STATE
		play $\act_{\run} \leftarrow \act_{\run+1}$
		\COMMENT{update action}
\ENDFOR
\end{algorithmic}

%% file: Figures/MU-MIMO-network.tex

\begin{tikzpicture}
[>=stealth,
vecstyle/.style = {->, double, line width=.5pt},
edgestyle/.style={-, line width=.5pt, black},
nodestyle/.style={circle, fill=Black,inner sep = .5pt}]

\footnotesize

\def\unit{5em}
\def\costhirty{0.8660256}
\def\cosfortyfive{0.7071068}


\coordinate (TX1) at (-2*\unit,.5);
\coordinate (RX1) at (1*\unit,0);

\coordinate (TX2) at (-3*\unit,-2.5*\unit);
\coordinate (RX2) at (-.5*\unit,-2*\unit);

\coordinate (TXK) at (3*\unit,-1*\unit);
\coordinate (RXK) at (2*\unit,-3*\unit);

\node (TX1) at (TX1) [label={[label distance=-3ex] 45:{TX$_{1}$}}] {\includegraphics[width=.5cm]{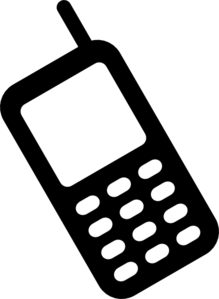}};
\node (RX1) at (RX1) [label={[label distance=-1ex] 45:{RX$_{1}$}}] {\includegraphics[width=.5cm]{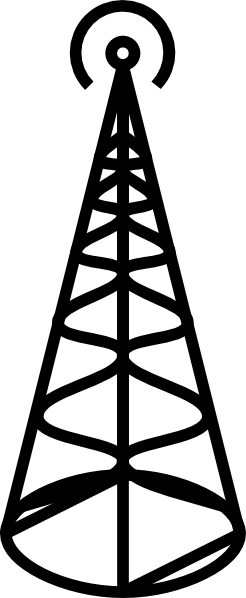}};

\node (TX2) at (TX2) [label={[label distance=-2ex] -60:{TX$_{2}$}}] {\includegraphics[width=.5cm]{Figures/phone.png}};
\node (RX2) at (RX2) [label={[label distance=-1ex] -50:{RX$_{2}$}}] {\includegraphics[width=.5cm]{Figures/base.png}};

\node (TXK) at (TXK) [label={[label distance=-3ex] 45:{TX$_{3}$}}] {\includegraphics[width=.5cm]{Figures/phone.png}};
\node (RXK) at (RXK) [label={[label distance=-1ex] -45:{RX$_{3}$}}] {\includegraphics[width=.5cm]{Figures/base.png}};

\draw [vecstyle,blue] (TX1) to node [midway, above, color=black] {$\bH_{11}$} (RX1);
\draw [vecstyle,blue] (TX2) to node [near end, above, color=black] {$\bH_{22}$} (RX2);
\draw [vecstyle,blue] (TXK) to node [midway, right, color=black] {\;$\bH_{33}$} (RXK);

\draw [vecstyle,dashed,red] (TX1) to node [midway, left, color=black] {$\bH_{12}$} (RX2);
\draw [vecstyle,dashed,red] (TX1) to node [midway, right, color=black] {\;$\bH_{13}$} (RXK);

\draw [vecstyle,dashed,red] (TX2) to node [near start, above, color=black] {$\bH_{21}$\;\;} (RX1);
\draw [vecstyle,dashed,red] (TX2) to node [midway, below, color=black] {$\bH_{23}$} (RXK);

\draw [vecstyle,dashed,red] (TXK) to node [near start, above, color=black] {$\bH_{31}$} (RX1);
\draw [vecstyle,dashed,red] (TXK) to node [midway, above, color=black] {$\bH_{32}$} (RX2);

\end{tikzpicture}

%% file: Algorithms.tex

We now turn to the fundamental underlying question arising from the definition of the regret:
\hilite{How can the optimizer attain a no-regret state, and what is the associated regret minimization rate?}

Of course, the answer to this question depends on the precise context of the problem at hand:
the geometry and dimensionality of the agent's action space,
the structure of the problem's objective (convex, strongly convex, etc.),
the information available to the optimizer (perfect gradient observations or otherwise),
are all factors that play a major role.
In the following, we focus on obtaining simple answers and identifying the resulting bottlenecks;
the various finer issues that arise (such as the impact of noise and feedback scarcity, etc.) are explored in \cref{sec:extras}.

\subsection{Online Gradient Descent}
\label{sec:OGD}

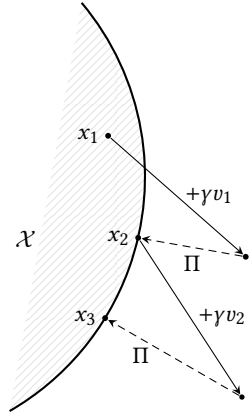
\begin{figure}[tbp]
\centering
\input{Figures/OGD.tex}
\caption{Schematic representation of \ac{OGD} (\cref{alg:OGD}).}
\label{fig:OGD}
\end{figure}

The most straightforward approach for solving classic, offline optimization problems is based on (projected) gradient descent:
at each stage, the algorithm takes a step against the gradient of the objective and, if necessary, projects back to the problem's feasible region.
Dating back to the seminal work of Zinkevich \cite{Zin03}, \acdef{OGD} is the direct adaptation of this idea to an online context.
In particular, writing
\begin{equation}
\payv_{\run}
	= -\nabla\loss_{\run}(\act_{\run})
\end{equation}
for the negative gradient of the $\run$-th loss function sampled at the agent's chosen action (again at round $\run$), \ac{OGD} can be described via the recursion
\begin{equation}
\label{eq:OGD}
\act_{\run+1}
	= \Eucl(\act_{\run} + \step\payv_{\run}),
\end{equation}
where
$\Eucl\from\R^{\vdim}\to\feas$ denotes the (Euclidean) projector
\begin{equation}
\label{eq:Eucl}
\Eucl(y)
	= \argmin_{\act\in\feas}\ \norm{y-\act}^{2},
\end{equation}
and $\step>0$ is a step-size parameter whose role is discussed below.
(For a schematic illustration and a pseudocode implementation, see \cref{fig:OGD} and \cref{alg:OGD} respectively.)


\begin{algorithm}[tbp]
\caption{\acf{OGD}}
\label{alg:OGD}
\input{Algorithms/OGD}
\end{algorithm}



\subsubsection{Basic regret guarantees}

The regret analysis of \ac{OGD} is remarkably simple and intuitive, so it is worth presenting in some detail.
Indeed, for any test point $\test\in\feas$, we have
\begin{equation}
\norm{\act_{\run+1} - \test}^{2}
	= \norm{\Eucl(\act_{\run} + \step\payv_{\run}) - \test}^{2}
	\leq \norm{\act_{\run} + \step\payv_{\run} - \test}^{2}
	= \norm{\act_{\run} - \test}^{2} + 2\step \product{\payv_{\run}}{(\act_{\run} - \test)} + \step^{2} \norm{\payv_{\run}}^{2}.
\end{equation}
Hence, after summing and rearranging, we obtain the ``linearized'' regret bound
\begin{equation}
\sum_{\run=1}^{\horizon} \product{\nabla\loss_{\run}(\act_{\run})}{(\act_{\run} - \test)}
	= -\sum_{\run=1}^{\horizon} \product{\payv_{\run}}{(\act_{\run} - \test)}
	\leq \frac{1}{2\step} \norm{\act_{1} - \test}^{2} + \frac{\step}{2} \sum_{\run=1}^{\horizon} \norm{\payv_{\run}}^{2}.
\end{equation}
With $\loss_{\run}$ convex, the first-order condition $\loss_{\run}(\act_{\run}) - \loss_{\run}(\test) \leq \product{\nabla\loss_{\run}(\act_{\run})}{(\act_{\run} - \test)}$ yields
\begin{equation}
\reg_{\horizon}
	\equiv \max_{\test\in\feas} \sum_{\run=1}^{\horizon} \bracks{\loss_{\run}(\act_{\run}) - \loss_{\run}(\test)}
	\leq \frac{\diamfeas^{2}}{2\step} + \frac{\step}{2} \sum_{\run=1}^{\horizon} \norm{\payv_{\run}}^{2},
\end{equation}
where $\diamfeas \equiv \max_{\act,\actalt\in\feas} \norm{\actalt - \act}$ denotes the (Euclidean) diameter of $\feas$.
Consequently, if we further assume that each $\loss_{\run}$ is $\Lip$-Lipschitz continuous (so $\norm{\payv_{\run}} \leq \Lip$ for all $\run$),%
\footnote{Recall here that $\obj\from\feas\to\R$ is $\Lip$-Lipschitz continuous if $\abs{\obj(\actalt) - \obj(\act)} \leq \Lip \norm{\actalt - \act}$ for all $\act,\actalt\in\feas$.}
we get
\begin{equation}
\reg_{\horizon}
	\leq \frac{\diamfeas^{2}}{2\step} + \frac{\step \Lip^{2}\horizon}{2}.
\end{equation}
Thus, optimizing the method's step-size $\step$ as a function of the horizon $\horizon$, we finally obtain:

\begin{theorem}[Worst-case regret of \ac{OGD}]
\label{thm:OGD}
Against $\Lip$-Lipschitz convex losses, the \ac{OGD} algorithm with step-size $\step = (\diamfeas/\Lip) /\sqrt{\horizon}$ enjoys the regret bound
\begin{equation}
\label{eq:reg-OGD}
\reg_{\horizon}
	\leq \diamfeas\Lip\sqrt{\horizon}.
\end{equation}
\end{theorem}

As in the case of adversarial bandits, this result is quite remarkable:
it shows that
\begin{inparaenum}
[\itshape a\upshape)]
\item
Cover's impossibility result does not extend to problems with a convex \textendash\ albeit possibly \emph{deterministic} \textendash\ structure;
and
\item
the passage from discrete to continuous action spaces does not worsen the achievable regret bounds in terms of the horizon of play.
\end{inparaenum}
In fact, \cref{thm:OGD} provides a key insight into what is needed to beat an adversary at its own game:
\emph{convexification.}
The use of mixed strategies (randomization) which played a crucial role in achieving no regret against an adversarial bandit can now be seen as an instance of \emph{convexifying} the problem's structure:
by looking at the mean (linear) rewards over the simplex of probability distributions (a convex set), the adversary can no longer trap the optimizer in a perpetual ``ping pong'' between extremes.
In essence, the existence of intermediate strategies (convex combinations of pure strategies) effectively mitigates the adversary's choices and provides the basis for achieving no regret.

\subsubsection{Minimax regret and refinements}

The guarantees of \cref{thm:OGD} cannot be improved in general:
indeed, since adversarial \aclp{MAB} fall under the umbrella of online linear optimization, it is not possible to achieve a worst-case guarantee better than $\Omega(\sqrt{\horizon})$ when facing linear losses.
More precisely, as was shown by Abernethy et al. \cite{ABRT08}, an informed adversary choosing linear losses of the form $\loss_{\run}(\act) = -{\payv_{\run}}{\act}$ with $\norm{\payv_{\run}} \leq \Lip$ can impose regret no less than
\begin{equation}
\label{eq:reg-lower-linear}
\reg_{\horizon}
	\geq \frac{\diamfeas\Lip}{2\sqrt{2}}\sqrt{\horizon}.
\end{equation}

The minimax bound \eqref{eq:reg-lower-linear} above suggests there is little hope of improving on the regret minimization rate of \ac{OGD} given by \eqref{eq:reg-OGD}.
Nevertheless, despite this negative result,
the optimizer can achieve significantly lower regret when facing \hilite{strongly convex} losses.
More precisely, if each $\loss_{\run}$ is $\strong$-strongly convex, Hazan et al. \cite{HAK07} showed that a slight modification of the \ac{OGD} algorithm%
\footnote{Specifically, the only difference between \cref{alg:OGD} and the variant of Hazan et al. \cite{HAK07} is the use of a variable step-size of the form $\step_{\run} = \strong/\run$ in the latter.}
enjoys the \emph{logarithmic} regret guarantee
\begin{equation}
\label{eq:reg-OGD-strong}
\reg_{\horizon}
	\leq \frac{1}{2} \frac{\Lip^{2}}{\strong} \log\horizon
	= \bigoh(\log\horizon).
\end{equation}
Remarkably, in the class of strongly convex functions, this guarantee is tight,
\emph{even up to the multiplicative constant in \eqref{eq:reg-OGD-strong}.}
In particular, Abernethy et al. showed in \cite{ABRT08} that if the adversary is restricted to quadratic convex functions of the form $\loss_{\run}(x) = \frac{1}{2}\act^{\top} \bM_{\run} \act - \product{\payv_{\run}}{\act} + c$ with $\bM_{\run} \mgeq \strong\,\bI$, the optimizer's worst-case regret cannot be better than
\begin{equation}
\reg_{\horizon}
	\geq \frac{1}{2}  \frac{\Lip^{2}}{\strong} \log\horizon.
\end{equation}

The above shows that the rate of regret minimization in online convex optimization depends crucially on the curvature of the loss functions encountered.
Against arbitrary loss functions, the optimizer cannot hope to do better than $\Omega(\sqrt{\horizon})$ in general;
however, if the optimizer's losses possess a strictly positive global curvature, the optimizer's worst-case guarantee drops down to $\bigoh(\log\horizon)$.
For convenience, we collect these guarantees in \cref{tab:reg-OGD}.


\begin{table}[tbp]
\centering
\renewcommand{\arraystretch}{1.2}
\input{Tables/Bounds-OGD}
\caption{Regret achieved by \ac{OGD} (\cref{alg:OGD}) against $\Lip$-Lipschitz convex losses.}
\label{tab:reg-OGD}
\end{table}

\begin{figure*}[t!]
\footnotesize
\subfigure[Energy efficiency under low mobility]{\label{fig:varying-EE-slow}%
\includegraphics[width=.48\textwidth]{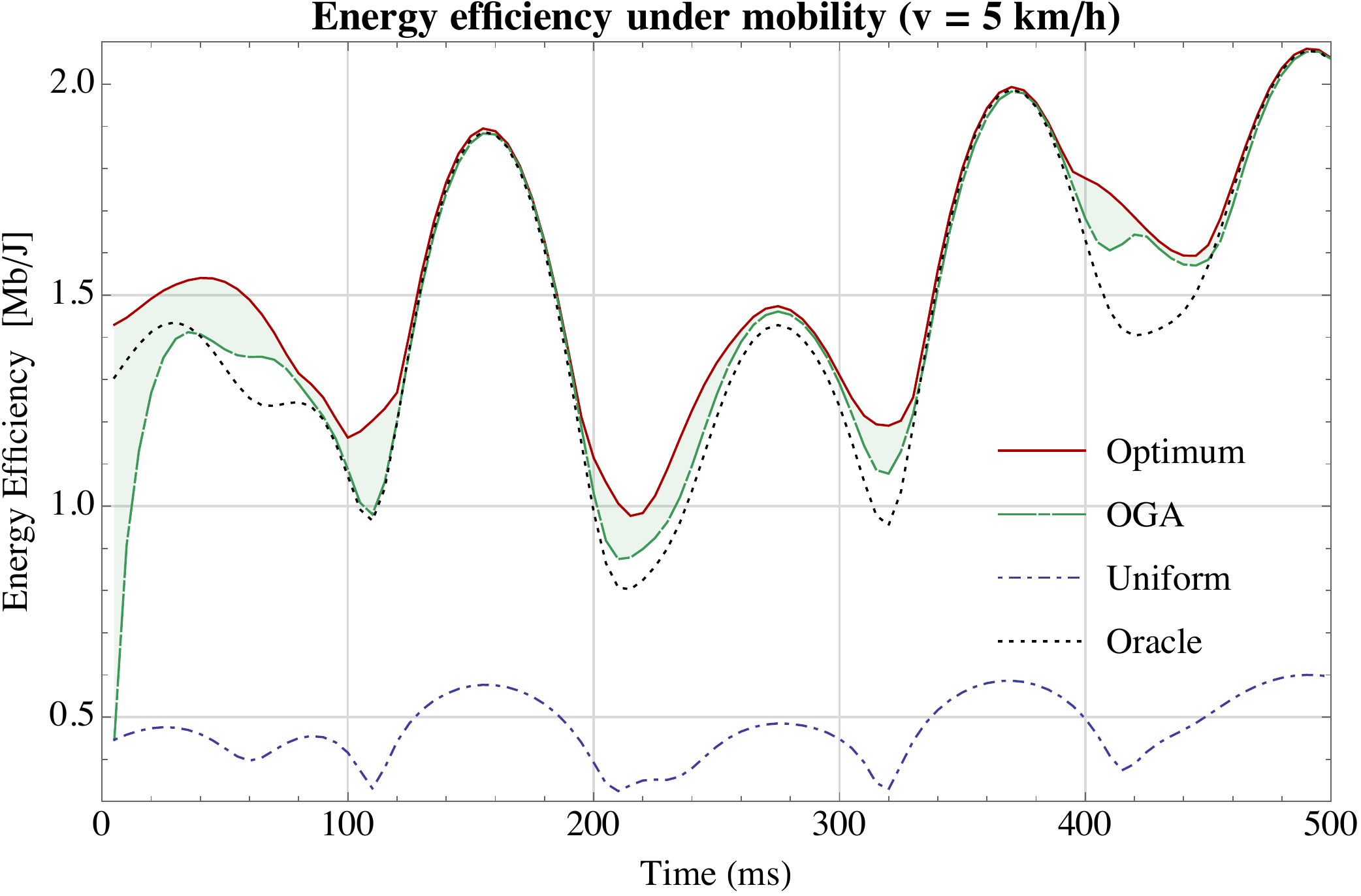}}
\hfill
\subfigure[Energy efficiency under high mobility]{\label{fig:varying-EE-fast}%
\includegraphics[width=.48\textwidth]{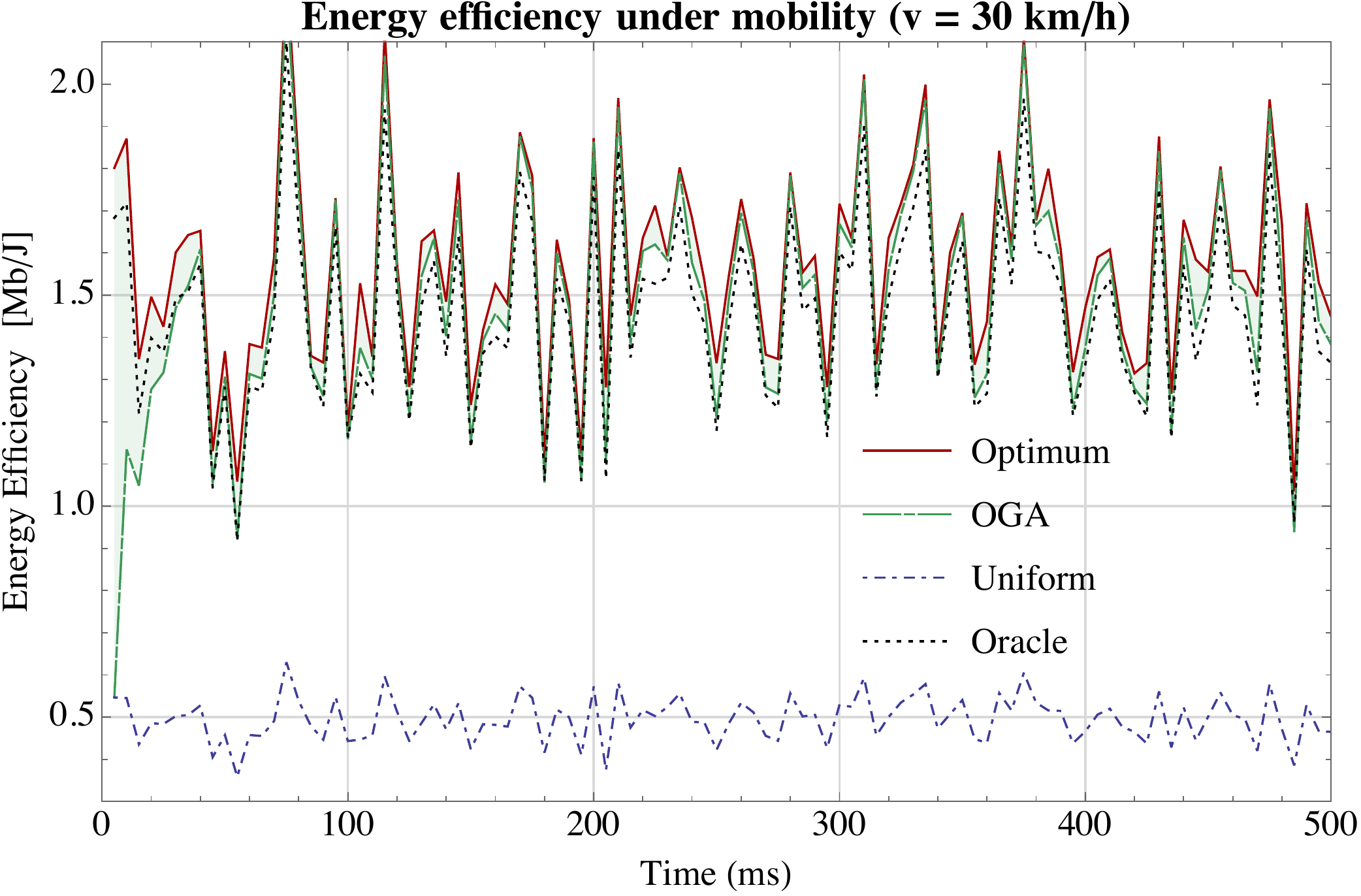}}
\\[.5ex]
\caption{Performance of an online gradient policy in a dynamic setting with mobile users seeking to optimize their transmit energy efficiency \cite{MB16}.
The users' achieved energy efficiency tracks its (evolving) maximum value remarkably well, even under rapidly changing channel conditions.}
\label{fig:OGD-MIMO-EE}
\end{figure*}

As an illustrative example, consider the maximization of transmit energy efficiency in multi-user \ac{MIMO} networks, defined here as the ratio between throughput and power consumption.
In \cite{MB16}, it was shown that if each user applies an online gradient policy,%
\footnote{Since users seek to \emph{maximize} their energy efficiency, the algorithm becomes an \emph{ascent} scheme, denoted as OGA in \cref{fig:OGD-MIMO-EE}.}
they are able to track their mean optimal transmit covariance matrix in a distributed manner. 
This is illustrated in \cref{fig:OGD-MIMO-EE} for mobile users with very different mobility characteristics, both pedestrian and vehicular.
\footnote{For complete details concerning the experimental setup (number of interfering users, device antennas, wireless characteristics, etc.), we refer the reader to \cite{MB16}.}
Specifically, in \cref{fig:OGD-MIMO-EE}, the users' policy is compared to several policies:
\begin{inparaenum}
[\itshape a\upshape)]
\item
the instantaneous optimum policy that varies over time (labeled ``Optimum'');
\item
a uniform power allocation scheme over antennas and frequency carriers (``Uniform'');
and
\item
the best fixed policy in hindsight (``Oracle'').
\end{inparaenum}
Remarkably, even under rapidly changing channel conditions, the users' online policy quickly converges to the maximum value and tracks its variations remarkably well and consistently outperforms the fixed oracle solution \textendash\ a behavior consistent with the no-regret property of \ac{OGD}.

\subsection{Online Mirror Descent}
\label{sec:OMD}

Even though the worst-case regret guarantee of \ac{OGD} is essentially tight, there is an important gap hidden in the problem's geometry.
This is best explained by going back to the adversarial \acl{MAB} problem:
as discussed in \cref{sec:setting}, \acp{MAB} can be seen as an online linear problem with feasible region $\feas = \simplex(\pures) = \setdef{\act\in\R^{\act}}{\sum_{\pure\in\pures} \act_{\pure} = 1}$ and linear loss functions of the form $\loss_{\run}(\act) = -\product{\payv_{\run}}{\act}$ where $\abs{\payv_{\pure,\run}} \leq 1$ for all $\pure\in\pures$ and all $\run=1,2,\dotsc,\horizon$.
This implies that the Lipschitz constant of the bandit's loss functions can be as high as
\begin{equation}
\label{eq:Lip-infty}
\txs
\Lip
	= \max\setdef*{\sqrt{\sum_{\pure\in\pures} \payv_{\pure}^{2}}}{\abs{\payv_{\pure}}\leq1}
	= \sqrt{1^{2} + \dotsm + 1^{2}}
	= \sqrt{\nPures},
\end{equation}
so the regret guarantee \eqref{eq:reg-OGD} of \ac{OGD} becomes
\begin{equation}
\reg_{\horizon}
	\leq \sqrt{\nPures\horizon}.
\end{equation}

Compared to the $\bigoh(\sqrt{\horizon \log\nPures})$ worst-case regret guarantee of the \ac{EW} algorithm (cf. \cref{sec:bandits}), the above indicates a gap of the order of $\bigoh(\sqrt{\nPures/\log\nPures})$ between the two algorithms \textendash\ and this, despite the fact that \ac{OGD} implicitly assumes full knowledge of the gradient vector $\payv_{\run}$ at each round.
In this way, we are faced with an apparent paradox:
\emph{how can the $\bigoh\parens{\sqrt{\horizon \log\nPures}}$ regret guarantee of the \ac{EW} algorithm be reconciled with the $\Omega\parens{\sqrt{\nPures\horizon}}$ minimax bound of \cref{eq:reg-lower-linear} for online linear problems with $\sqrt{\nPures}$-Lipschitz losses?}

The answer to this paradox lies in the different sets of the reward vectors $\payv_{\run}$ chosen by the adversary.
In online linear problems with $\sqrt{\nPures}$-Lipschitz losses, $\payv_{\run}$ is drawn from the Euclidean ball of radius $\sqrt{\nPures}$ (\ie $\norm{ \payv_{\run}} \leq \sqrt{\nPures}$).
By contrast, in \acl{MAB} problems, the reward vector $\payv_{\run}$ lies in the hypercube $[-1,1]^{\nPures}$ (since $\abs{\payv_{\pure,\run}} \leq 1$).
The hypercube is strictly contained in the (much) larger Euclidean ball so, heuristically, an adversarial \acl{MAB} is more constrained and cannot cause as much damage.

%
%

Geometrically, this highlights a crucial difference between the $\ell^{\infty}$ sup-norm and the ordinary (Euclidean) $\ell^{2}$ norm, which in turn explains the performance gap between the \ac{EW} and \ac{OGD} algorithms in online linear problems.%
\footnote{Recall here that the hypercube $[-1,1]^{\nPures}$ is simply the $\ell^{\infty}$ unit ball, \ie $\norm{\payv}_{\infty} \equiv \max_{\pure\in\nPures} \payv_{\pure} \leq 1$.}
By moving beyond the Euclidean norm and geometry might bridge this gap and lift the so-called \hilite{curse of dimensionality} in general online convex optimization problems.
Especially in the world of Big Data, many problems of practical interest have state spaces of huge dimension, so descent methods must be finely tuned to the problem's geometry in order to attain an optimal regret minimization rate.


\begin{algorithm}[tbp]
\caption{\acf{OMD}}
\label{alg:OMD}
\input{Algorithms/OMD}
\end{algorithm}


\subsubsection{Mirror descent}

A systematic way to exploit the geometry of the problem is via the method of \acdef{OMD} \cite{SS07}.
To illustrate the core components of the method (which can be traced back to the seminal work of Nemirovski and Yudin \cite{NY83} for offline problems), it is convenient to rewrite the Euclidean update step \eqref{eq:OGD} of \ac{OGD} as
\begin{flalign}
\act_{\run+1}
	= \Eucl(\act_{\run} + \step\payv_{\run})
	&= \argmin_{\act\in\feas} \ \tfrac{1}{2}\norm{\act_{\run} + \step\payv_{\run} - \act}^{2}
	\notag\\
	&= \argmin_{\act\in\feas} \ \braces[\big]{\tfrac{1}{2}\norm{\act_{\run} - \act}^{2}
		+ \tfrac{1}{2}\norm{\step\payv_{\run}}^{2}
		+ \step\product{\payv_{\run}}{(\act_{\run} - \act)}}
	\notag\\
	&= \argmin_{\act\in\feas} \ \braces[\big]{\step\product{\payv_{\run}}{(\act_{\run} - \act)} + \breg(\act,\act_{\run})},
\end{flalign}
where we have defined
\begin{equation}
\txs
\breg(\base,\act)
	= \frac{1}{2} \norm{\base - \act}^{2}
	= \frac{1}{2} \norm{\base}^{2} - \frac{1}{2} \norm{\act}^{2} - \product{\act}{(\base - \act)}.
\end{equation}
The key novelty of \acl{MD} is to replace this quadratic term by the so-called \emph{Bregman divergence}
\begin{equation}
\label{eq:Bregman}
\breg_{h}(\base,\act)
	= h(\base) - h(\act) - \product{\nabla h(\act)}{(\base - \act)},
\end{equation}
where $h\from\feas\to\R$ is a smooth $K$-strongly convex function (usually referred to as a \emph{regularizer}).
In so doing, we obtain the \acdef{OMD} algorithm
\begin{equation}
\label{eq:OMD}
\act_{\run+1}
	= \prox_{\act_{\run}}(\step_{\run}\payv_{\run}),
\end{equation}
where the \emph{mirror-prox operator} $\prox$ is defined as
\begin{equation}
\label{eq:mirror}
\prox_{\act}(\payv)
	= \argmin_{\actalt\in\feas} \ \braces*{\product{\payv}{(\act-\actalt)} + \breg_{h}(\actalt,\act)},
\end{equation}
and, as before, $\payv_{\run} = -\nabla\loss_{\run}(\act_{\run})$ denotes the negative gradient of the loss function of the $\run$-th sampled at $\act_{\run}$.
(For a pseudocode implementation, see \cref{alg:OMD}.)

\subsubsection{Examples}
\label{sec:OMD_examples}

Before delving into the regret guarantees of the \ac{OMD} algorithm, it is worth discussing some important examples in detail:

\setcounter{example}{0}
\begin{example}[Euclidean regularization]
As we discussed above, the regularizer $h(\act) = \frac{1}{2}\norm{\act}^{2}$ yields the archetypal \ac{OGD} algorithm \eqref{eq:OGD}.
As such, all results obtained for \ac{OMD} also carry over to \ac{OGD}.
\end{example}

\begin{example}[Entropic regularization]
Another important instance of the \ac{OMD} method is when the problem's feasible region $\feas$ is the unit simplex of $\R^{\vdim}$ and the regularizer is the (negative) Gibbs\textendash Shannon entropy
\begin{equation}
\label{eq:entropy}
h(\act)
	= \sum_{j=1}^{\vdim} \act_{j} \log \act_{j}.
\end{equation}
A short calculation shows that the resulting mirror-prox operator is given by the exponential mapping
\begin{equation}
\prox_{\act}(\payv)
	= \frac{(\act_{j}\exp(\payv_{j}))_{j=1}^{\vdim}}{\sum_{k=1}^{\vdim} \act_{k}\exp(\payv_{k})}
\end{equation}
which in turn leads to the \acdef{EG} algorithm:
\begin{equation}
\act_{j,\run+1}
	= \frac{\act_{j,\run} \exp(\step \payv_{j,\run})}{\sum_{k=1}^{\vdim} \act_{k,\run}  \exp(\step\payv_{k,\run})}.
\end{equation}
Importantly, going back to the \ac{EW} algorithm \eqref{eq:EW} for \aclp{MAB}, we readily get
\begin{equation}
\act_{j,\run+1}
	\propto \exp(\score_{j,\run+1})
	= \exp(\score_{j,\run} + \step\payv_{j,\run})
	\propto \act_{j,\run}\exp(\step \payv_{j,\run}).
\end{equation}
Thus, by normalizing, we see that
\ac{OMD} with entropic regularization is simply the \ac{EW} algorithm.
This remarkable observation sheds further light on the structure of the \ac{EW} algorithm:
despite their very different origins, \acl{EW} and gradient descent are different sides of the same coin \textendash\
\hilite{mirror descent.}
\end{example}


\begin{algorithm}[tbp]
\caption{\acf{MXL}}
\label{alg:MXL}
\input{Algorithms/MXL}
\end{algorithm}

\begin{figure*}[tbp]
\footnotesize
\subfigure[Sum rate for 20 users and 8 receive antennas]{\label{fig:varying-EE-slow}%
\includegraphics[width=.48\textwidth]{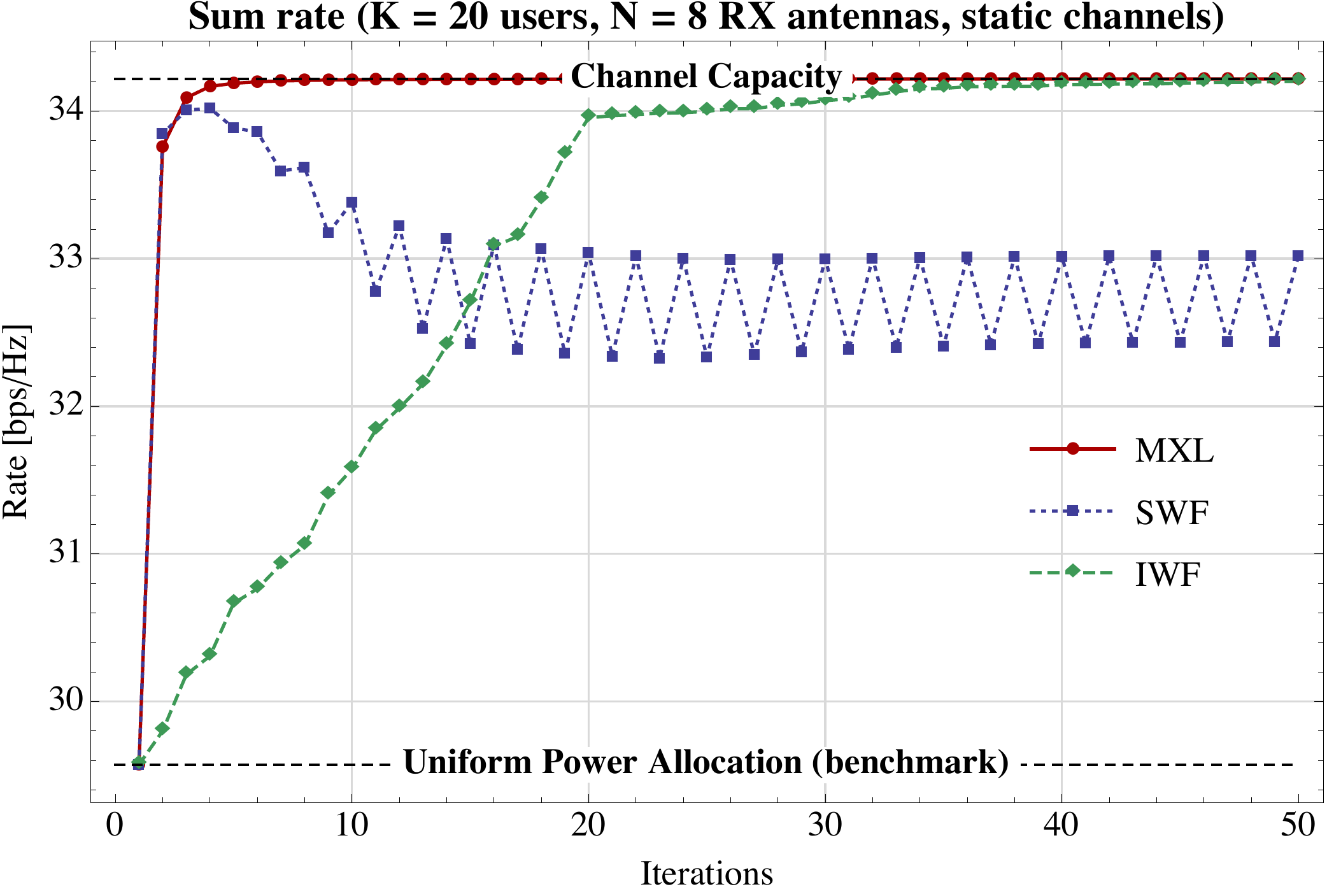}}
\hfill
\subfigure[Sum rate for 50 users and 32 receive antennas]{\label{fig:varying-EE-fast}%
\includegraphics[width=.48\textwidth]{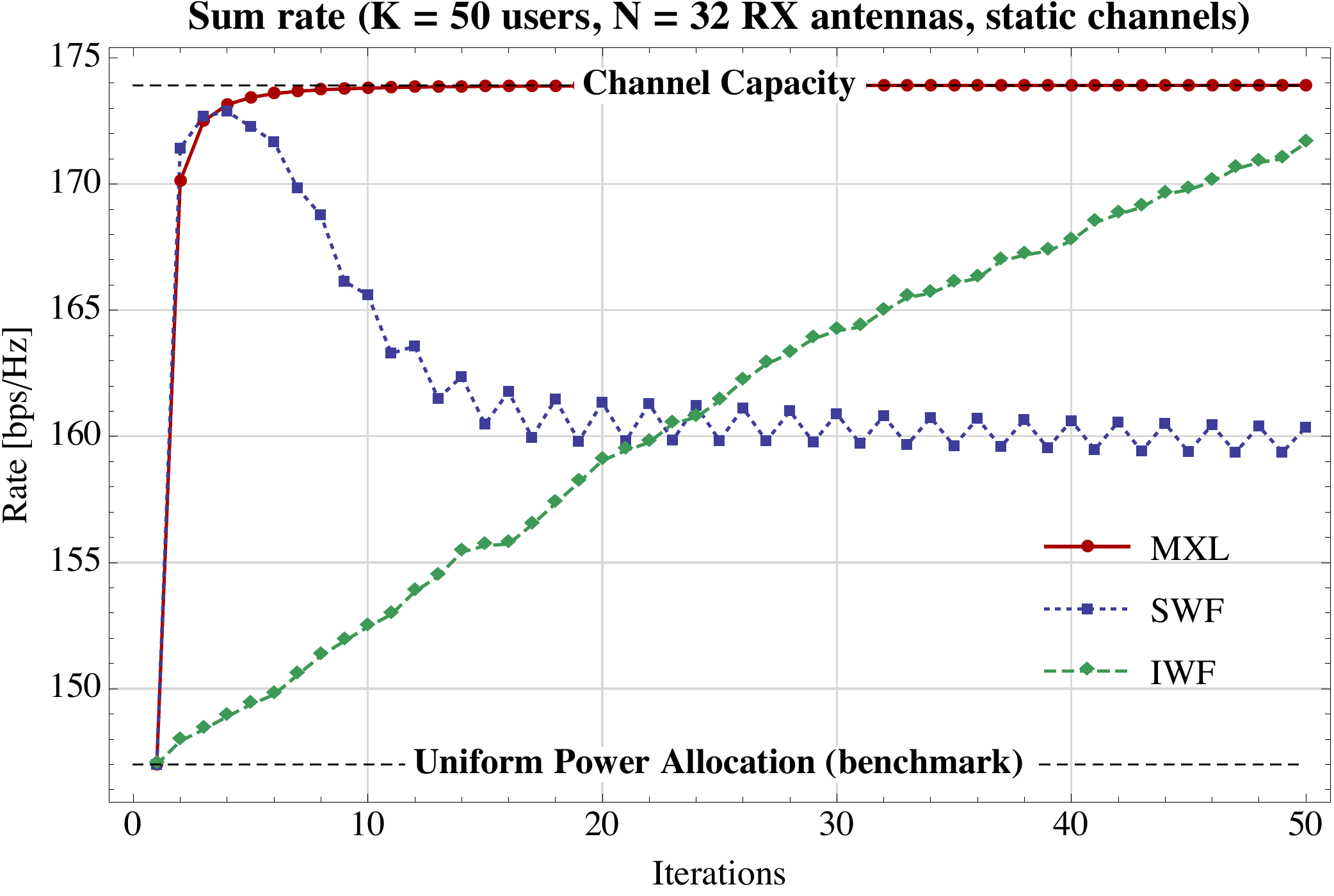}}
\\[1ex]
\caption{Performance comparison between the \ac{MXL} algorithm with water-filling methods in a static uplink setting with one base station and multiple transmitters driven by their own rates \cite{MM16}.
\ac{IWF} converges slowly because only one user updates its input covariance per iteration;
\ac{SWF} is faster but does not always converge due to cycles in the update process.
On the contrary, the \ac{MXL} algorithm converges within a few iterations even for large numbers of users.}
\label{fig:MIMO-rates}
\end{figure*}

\begin{example}[Matrix exponential learning]
Our last example concerns learning with matrices and is of particular relevance to \ac{MIMO} communications \cite{SPFP10,SP10,MM16,MBNS17}.
When optimizing signal covariance matrices in such systems, the problem's feasible region is often a spectrahedron of the form
\begin{equation}
\label{eq:spectron}
\boldsymbol{\feas}
	= \setdef{\bX\in\R^{\vdim\times\vdim}}{\bX\mgeq0,\,\tr{\bX} \leq 1}.
\end{equation}
A tailor-made regularizer for this type of constraints is given by the (negative) \emph{von Neumann entropy}
\begin{equation}
\label{eq:quantum}
h(\bX)
	= \trof{\bX\log\bX} + (1 - \tr\bX) \log(1 - \tr\bX).
\end{equation}
As was shown in \cite{MM16,MBNS17}, this choice of regularizer yields the \acdef{MXL} algorithm
\begin{equation}
\begin{aligned}
\bY_{\run+1}
	&= \bY_{\run} + \step\bV_{\run}
	\\
\bX_{\run+1}
	&= \frac{\exp(\bY_{\run+1})}{1+ \trof{ \exp(\bY_{\run+1})}} \label{eq:MXL}
\end{aligned}
\end{equation}
where $\bV_{\run} = -\nabla_{\bX_{\run}}\loss_{\run}(\bX_{\run})$ is the matrix gradient of the $\run$-th round loss function (for a pseudocode implementation, see \cref{alg:MXL}).%
\footnote{Since $\loss_{\run}$ is a real function of a Hermitian matrix, its gradient is also a Hermitian matrix, so $\exp(\bY_{\run})$ is positive-definite.}
A very appealing property of the exponential update in \eqref{eq:MXL} is that it ensures positive-definiteness in an elegant and lightweight manner compared to the Euclidean projection on the positive-definite cone (which requires solving a convex optimization problem at each stage).
Moreover, this algorithm exhibits very fast convergence rates for throughput and energy efficiency maximization in wireless networks, greatly outperforming traditional algorithms based on water-filling \cite{MM16,MBNS17};
for an illustration of the algorithm's performance in \ac{MIMO} systems, see \cref{fig:MIMO-rates}.%
\footnote{The network parameters and numerical setup are fully described in \cite{MM16}.}
\end{example}


\begin{figure*}[t!]
\footnotesize
\subfigure[Classification errors on Iris, Wine and MNIST.]{\label{fig:test-errors}%
\includegraphics[width=.38\textwidth]{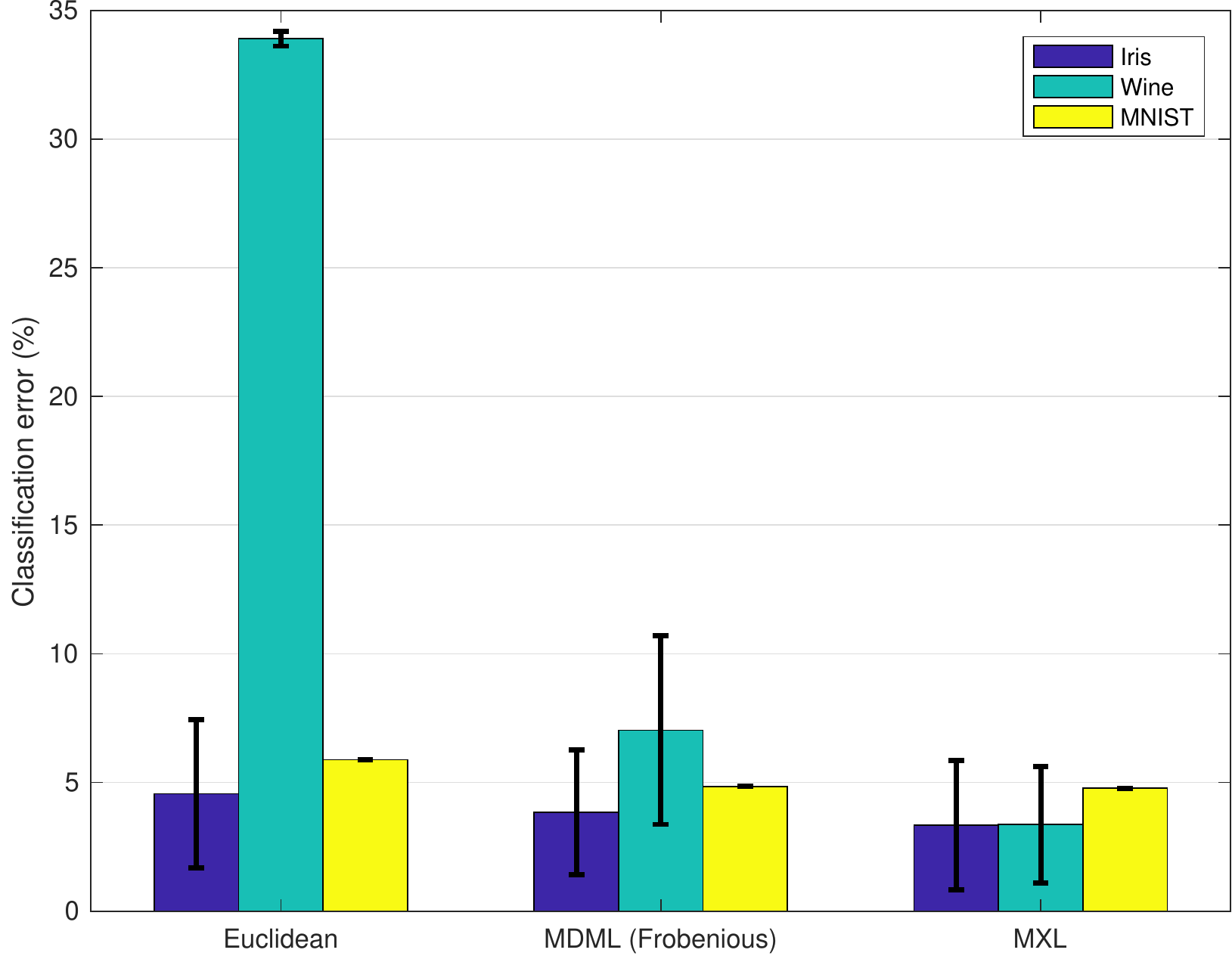}}
\hfill
\subfigure[Principal data components in the Euclidean and the transformed space via \ac{MXL} on the Wine dataset.]{\label{fig:wine-mxl}%
\includegraphics[width=.58\textwidth]{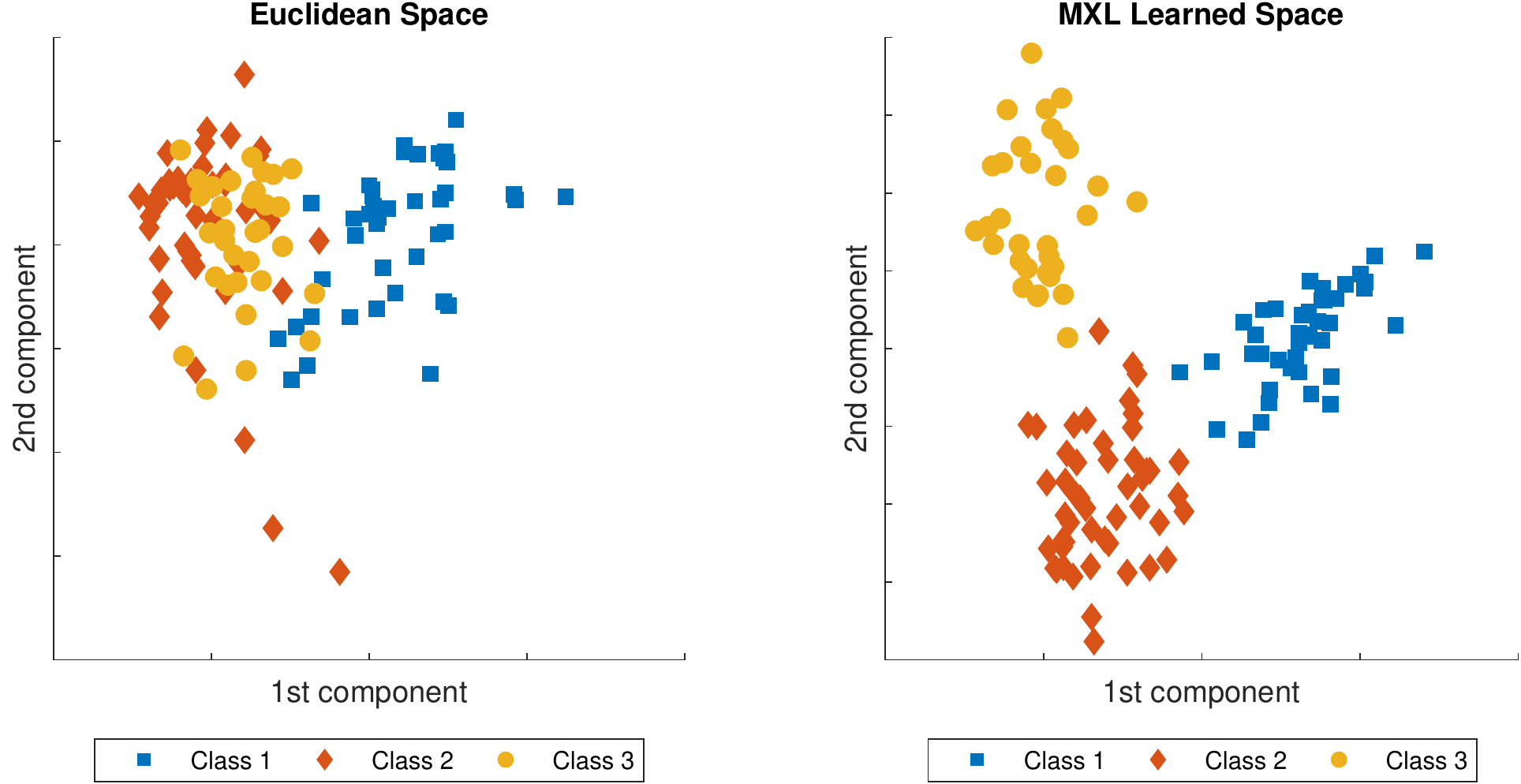}}
\\[.5ex]
\caption{Performance comparison between a Euclidean metric, and two online metrics based on mirror descent: the MDML in \cite{Kunapuli2012} and \ac{MXL}, on three different datasets (Iris, Wild and MNIST). Both online metrics always outperform the Euclidean metric and the \ac{MXL} algorithm provides the best online learned metric. In the Wine dataset, in which the Euclidean representation is quite ill-suited for classification, the \ac{MXL} algorithm provides the best linear transformation, which successfully separates the images into the three classes.}
\label{fig:OnlineMetricExp}
\end{figure*}

To highlight the range of the \ac{MXL} algorithm, we also consider below its application to the multimedia indexing problem described in \cref{sec:setting}.
Specifically, in \cref{fig:OnlineMetricExp}, we compare the Euclidean metric with the online metric that was learned by the \ac{MXL} algorithm, and a different online metric based on the \ac{MDML} algorithm of \cite{Kunapuli2012}.
In more detail, we tested the \ac{MDML} variant based on the Frobenius norm regularization (as opposed to the entropic regularization).
\cref{fig:OnlineMetricExp} illustrates that \ac{MXL} performs best in terms of classification test errors on three well-known datasets:
Iris, Wine and MNIST.%
\footnote{Datasets available at \href{https://archive.ics.uci.edu/ml/datasets/iris}{https://archive.ics.uci.edu/ml/datasets/iris}, \href{https://archive.ics.uci.edu/ml/datasets/wine}{https://archive.ics.uci.edu/ml/datasets/wine}, and \href{http://yann.lecun.com/exdb/mnist/}{http://yann.lecun.com/exdb/mnist/} respectively.}
both the mean and variance of the results with respect to the random dataset splitting between train and test over $200$ runs are plotted (except for the MNIST data in which the pre-defined sets were used).
For a detailed description of the experimental setup used to obtain these plots (dataset description, choice of the train and test datasets, step-sizes, etc.) we refer the reader to the online report \cite{BMNS_techrep2018}.


\medskip
\subsubsection{Regret guarantees}

The basic worst-case guarantee of \ac{OMD} is as follows \cite{SS11,BCB12,KM17}:

\begin{theorem}[Worst-case regret of \ac{OMD}]
\label{thm:OMD}
Against $\Lip$-Lipschitz convex losses, the \ac{OMD} algorithm based on a $K$-strongly convex regularizer $h$ enjoys the regret bound
\begin{equation}
\label{eq:reg-OMD}
\reg_{\horizon}
	\leq 2\Lip\sqrt{\frac{\max h - \min h}{2K} \horizon},
\end{equation}
achieved by taking the step-size $\step = \Lip^{-1} \sqrt{2K(\max h - \min h)/\horizon}$.
\end{theorem}

An important remark in the above is that the strong convexity and Lipschitz constants $K$ and $\Lip$ need not be taken with respect to the Euclidean norm.%
\footnote{However, both constants must be tied to norms that are ``compatible'' (dual to each other) \cite{SS11}.}
When they are, \cref{thm:OMD} readily yields the $\bigoh(\Lip\sqrt{\horizon})$ regret guarantee of \ac{OGD}.%
\footnote{In fact, by using the ``centered'' regularizer $h(\act) = \frac{1}{2} \norm{\act - \act_{c}}^{2}$ where $\act_{c}$ is the center of the smallest ball enclosing $\feas$, the bound \eqref{eq:reg-OMD} shaves off an additional factor of $2$ from the bound \eqref{eq:reg-OGD} \cite{KM17}.}
For the \ac{EW} algorithm, the entropic regularizer \eqref{eq:entropy} is $1$-strongly convex with respect to the taxicab ($\ell^{1}$) norm \cite{SS11} and $\max h - \min h = \sum_{j=1}^{\vdim} \frac{1}{\vdim} \log\vdim = \log\vdim$, leading to the $\bigoh\parens{\sqrt{\horizon\log\nPures}}$ bound \eqref{eq:reg-EW} (since $\nPures \equiv \vdim$).
Finally, for the \ac{MXL} algorithm, the von Neumann regularizer $h$ is $1$-strongly convex relative to the nuclear matrix norm ($\norm{\bX}_{1} = \tr\abs{\bX}$) and $\max h - \min h = \log\vdim$, leading to the regret bound
\begin{equation}
\reg_{\horizon}
	\leq \Lip\sqrt{2\horizon \log\vdim}.
\end{equation}
Remarkably, the regret bound of the \ac{EW} and \ac{MXL} algorithms coincide, even though the latter is run in a space of much greater dimension ($\vdim\times\vdim$ versus $\vdim$ respectively).

We summarize these results in \cref{tab:reg-OMD}.
The main take-away from this table is that \ac{OMD} enjoys the same $\sqrt{\horizon}$ rate as \ac{OGD}, but the multiplicative constants are optimized relative to the dimension and geometry of the problem.
This logarithmic reduction is of immense value to real-world Big Data problems that suffer from the curse of dimensionality.
As a result, the systematic design of tailor-made \ac{OMD} algorithms for arbitrary problem geometries has attracted considerable interest in the literature and remains a vigorously researched question, often resembling a form of art.


\begin{table}[tbp]
\centering
\renewcommand{\arraystretch}{1.2}
\input{Tables/Bounds-OMD}
\caption{Regret achieved by \ac{OMD} strategies in various geometries.}
\label{tab:reg-OMD}
\end{table}


%% file: Algorithms/OGD.tex

\begin{algorithmic}[1]
\REQUIRE
	step-size $\step>0$
\STATE
	choose $\act\in\feas$
	\COMMENT{initialization}%
\FOR{$\run=1$ \TO $\horizon$}
	\STATE
		incur loss $\hat\loss \leftarrow \loss_{\run}(\act)$
		\COMMENT{losses revealed}%
	\STATE
		observe $\payv \leftarrow - \nabla\loss_{\run}(\act)$
		\COMMENT{gradient feedback}%
	\STATE
		play $\act \leftarrow \Eucl(\act + \step\payv)$
		\COMMENT{gradient step}%
\ENDFOR
\end{algorithmic}

%% file: Figures/OGD.tex

\colorlet{TangentColor}{blue}
\colorlet{PolarColor}{red}

\begin{tikzpicture}
[scale=1.2,
nodestyle/.style = {circle,fill=black,inner sep=0, minimum size=2}]

\small

\draw [thick,pattern = north east lines, pattern color = black!10] (45:3) arc (40:-60:3);
\draw node at (1.5,-.5) {$\feas$};

\node [nodestyle] (x1) at (15:2.5) {.};
\node [left] at (x1) {$\act_{1}$};

\node [nodestyle] (y1) at (-10:4) {.};

\node [nodestyle] (x2) at (-10:2.79) {.};
\node [left] at (x2) {$\act_{2}$};

\node [nodestyle] (y2) at (-30:4.5) {};

\node [nodestyle] (x3) at (-30:2.75) {.};
\node[left] at (x3) {$\act_{3}$};

\draw[-stealth] (x1) -- (y1) node [midway, right]{$+\step\payv_{1}$};
\draw [-stealth, densely dashed] (y1) -- (x2) node [midway,below] {$\Eucl$};
\draw [-stealth] (x2) -- (y2) node[midway, right]{$+\step\payv_{2}$};
\draw[-stealth, densely dashed] (y2) -- (x3) node [near end,below] {$\Eucl$};

\end{tikzpicture}

%% file: Tables/Bounds-OGD.tex

\small
\begin{tabular}{rll}
\;
	&\scshape
	Minimax Regret
	&\scshape
	Worst-Case
	\\
\hline
\scshape
Convex Losses
	&$\Omega\parens{\Lip\horizon^{1/2}}$
	&$\bigoh\parens{\Lip\horizon^{1/2}}$
	\\
\hline
\scshape
Linear Losses
	&$\Omega\parens{\Lip\horizon^{1/2}}$
	&$\bigoh\parens{\Lip\horizon^{1/2}}$
	\\
\hline
\scshape
$\strong$-Strongly Convex
	&$\Omega\parens{\strong^{-1}\Lip^{2} \log\horizon}$
	&$\bigoh\parens{\strong^{-1}\Lip^{2} \log\horizon}$
	\\
\hline
\end{tabular}
\vspace{2ex}

%% file: Algorithms/OMD.tex

\begin{algorithmic}[1]
\REQUIRE
	$K$-strongly convex regularizer $h\from\feas\to\R$,
	step-size $\step>0$
\STATE
	choose $\act\in\feas$
	\COMMENT{initialization}%
\FOR{$\run=1$ \TO $\horizon$}
	\STATE
		incur loss $\hat\loss \leftarrow \loss_{\run}(\act)$
		\COMMENT{losses revealed}%
	\STATE
		observe $\payv \leftarrow - \nabla\loss_{\run}(\act)$
		\COMMENT{gradient feedback}%
	\STATE
		play $\act \leftarrow \prox_{\act}(\step\payv)$
		\COMMENT{mirror step}%
\ENDFOR
\end{algorithmic}

%% file: Algorithms/MXL.tex

\begin{algorithmic}[1]
\REQUIRE
	step-size $\step>0$
\STATE
	$\bY\leftarrow0$
	\COMMENT{initialization}%
\FOR{$\run=1$ \TO $\horizon$}
	\STATE
		set $\bX \leftarrow \exp(\bY) / [1 + \tr(\exp(\bY))]$
		\COMMENT{matrix exponentiation}%
	\STATE
		incur loss $\hat\loss \leftarrow \loss_{\run}(\bX)$
		\COMMENT{losses revealed}%
	\STATE
		observe $\bV \leftarrow -\nabla\loss_{\run}(\bX)$
		\COMMENT{gradient feedback}%
\ENDFOR
\end{algorithmic}

%% file: Tables/Bounds-OMD.tex

\small
\begin{tabular}{rlll}
\;
	&\scshape
	Region
	&\scshape
	Losses
	&\scshape
	Regret
	\\
\hline
\scshape
\ac{OGD}
	&\scshape
	Unit ball
	&$\norm{\payv_{\run}} \leq 1$
	&$\bigoh\parens{\sqrt{\horizon}}$
	\\
\hline
\scshape
\ac{EW}
	&\scshape
	Simplex
	&$\norm{\payv_{\run}}_{\infty} \leq 1$
	&$\bigoh(\sqrt{\horizon\log\vdim})$
	\\
\hline
\scshape
\ac{MXL}
	&\scshape
	Spectrahedron
	&$\abs{\eig(\bV_{\run})} \leq 1$
	&$\bigoh(\sqrt{\horizon\log\vdim})$
	\\
\hline
\end{tabular}
\vspace{2ex}

%% file: Extras.tex

In this final section, we discuss some further aspects of regret minimization that are of particular relevance to signal processing and its applications.

\subsection{Imperfect feedback}
\label{sec:imperfect}

We begin by discussing the quality and amount of feedback available to the optimizer.
Indeed, throughout the analysis of the previous section, it was tacitly assumed that the optimizer had access to a black-box feedback mechanism \textendash\ an \emph{oracle} \textendash\ which, when called, provided \emph{perfect gradient observations} for the loss function of each stage.
In the \acl{MAB} language of \cref{sec:bandits}, this corresponds to the ``full information'' case, \ie when the agent has full knowledge of the payoff vector $\payv_{\run}$ at each stage (meaning the payoff of each arm \textendash\ chosen or not).
In practical situations (\eg throughput maximization in a fast-fading wireless medium), such information might be difficult \textendash\ or even impossible \textendash\ to come by, so the relevance and applicability of the online algorithms we have discussed so far is put to the test.

To model imperfections in the optimizer's feedback, we will focus on an application-agnostic framework where, at each stage $\run=1,2,\dotsc$, the optimizer receives a noisy, imperfect estimate $\est_{\run}$ of the true gradient $\payv_{\run}= -\nabla\loss_{\run}(\act_{\run})$.
For this estimate, we make the standard statistical assumptions that it is
\emph{unbiased and bounded in mean square},
\ie
\begin{subequations}
\label{eq:estimate}
\begin{flalign}
\label{eq:unbiased}
&\exof{\est_{\run} \given \filter_{\run-1}}
	= \payv(\act_{\run}),
\intertext{and}
\label{eq:vbound}
&\exof{ \norm{\est_{\run}}^{2} \given \filter_{\run-1}}
	\leq \vbound^{2}
	\quad
	\text{for some finite $\vbound>0$},
\end{flalign}
\end{subequations}
where $\filter_{\run}$ denotes the history of play up to stage $\run$ (inclusive).
These bare-bones assumptions simply indicate that there is no systematic error in the optimizer's gradient measurements and that the observed error cannot become unreasonably large with high probability.
This is true for most statistical distributions used to describe error processes in practice \textendash\ ranging from uniform and (sub-)Gaussian to log-normal, exponential, and Lévy-type models \textendash\ so we will maintain this assumption throughout what follows.

Of course, given that the feedback to the optimizer is now a stochastic process, the incurred regret will also be random.
On that account, the figure of merit will now be the optimizer's mean regret
\begin{equation}
\label{eq:reg-mean-opt}
\bar\reg_{\horizon}
	= \max_{\act\in\feas} \sum_{\run=1}^{\horizon} \exof{\loss_{\run}(\act_{\run}) - \loss_{\run}(\act)},
\end{equation}
which is the direct extension of the corresponding regret measure \eqref{eq:reg-mean-adv} for \acl{MAB} problems (\cf \cref{sec:bandits}).
Then, focusing for simplicity on the \ac{OGD} algorithm, we have the following worst-case guarantee \cite{BCB12,KM17}:

\begin{theorem}[Worst-case regret of \ac{OGD} with imperfect feedback]
\label{thm:OGD-noisy}
Against convex losses, the \ac{OGD} algorithm with
step-size $\step = \diamfeas/\sqrt{\vbound^{2}\horizon}$
and
noisy gradient feedback satisfying \eqref{eq:estimate}
enjoys the worst-case guarantee
\begin{equation}
\label{eq:reg-OGD-noisy}
\bar\reg_{\horizon}
	\leq \diamfeas\vbound\sqrt{\horizon}
	=\bigoh(\sqrt{\horizon}).
\end{equation}
\end{theorem}

Remarkably, despite all the uncertainty, the worst-case regret of \ac{OGD} with noisy feedback is of the same order in $\horizon$ as in the perfect information case (\cf \cref{thm:OGD} in \cref{sec:algorithms}).
That being said, the noise \emph{does} deteriorate the algorithm's performance in two important ways:
\begin{enumerate}
\item
The regret bound \eqref{eq:reg-OGD-noisy} depends linearly on the second moment of the optimizer's feedback (which could become quite high if the optimizer only has access to low-quality observations with a lot of noise).
\item
The optimizer must be able to estimate the variance of the noise in order to run \ac{OGD}.
\end{enumerate}
Especially the latter could have important ramifications for the agent's regret:
if the optimizer ``plays it safe'' and overestimates the variance of the noise, the incurred regret will grow commensurately because of the second moment factor $\vbound$ in \eqref{eq:reg-OGD-noisy}.
This quantifies the impact of errors and observation noise in the performance of \ac{OGD}.


\begin{figure*}[t]
\includegraphics[width=.48\textwidth]{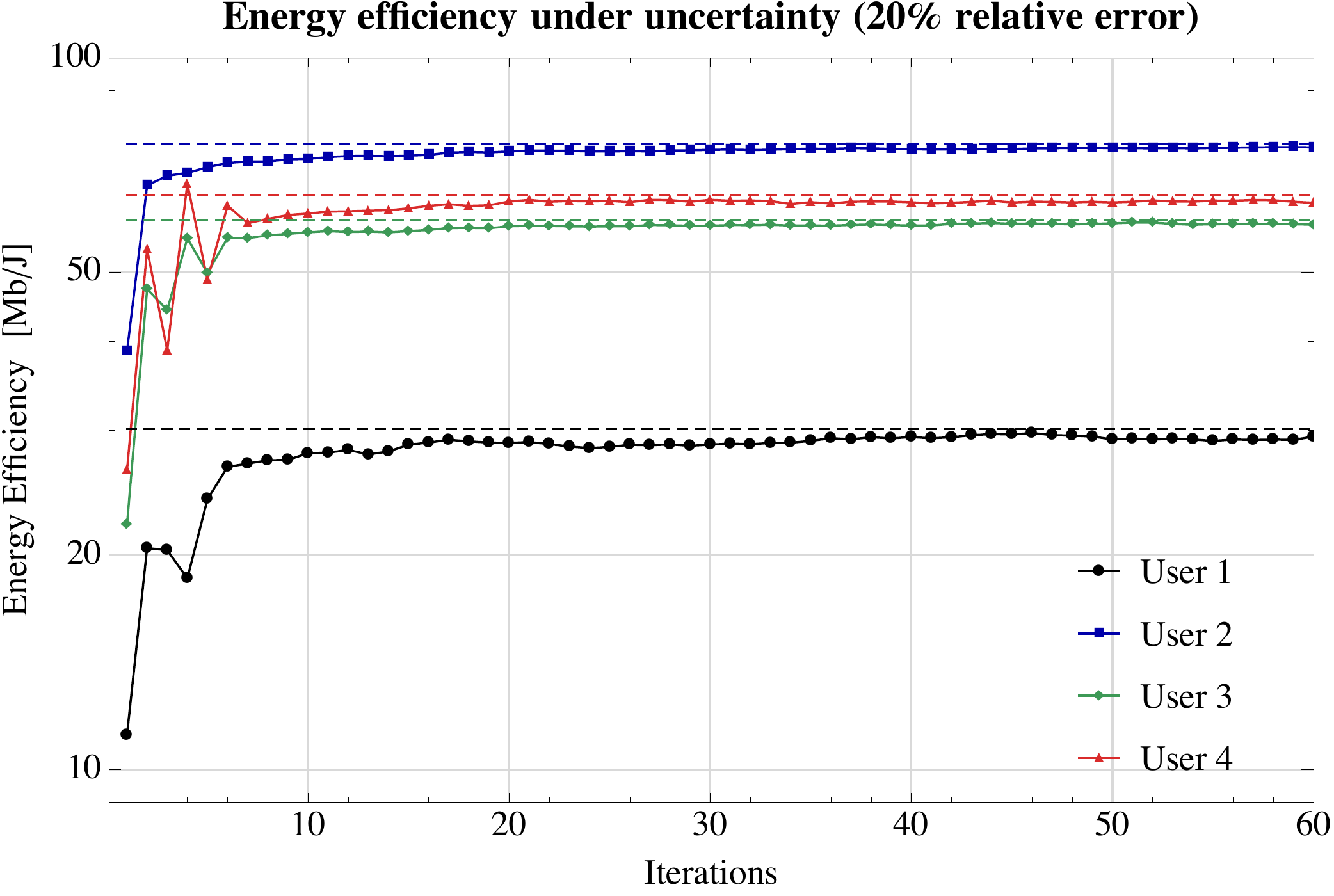}
\hfill
\includegraphics[width=.48\textwidth]{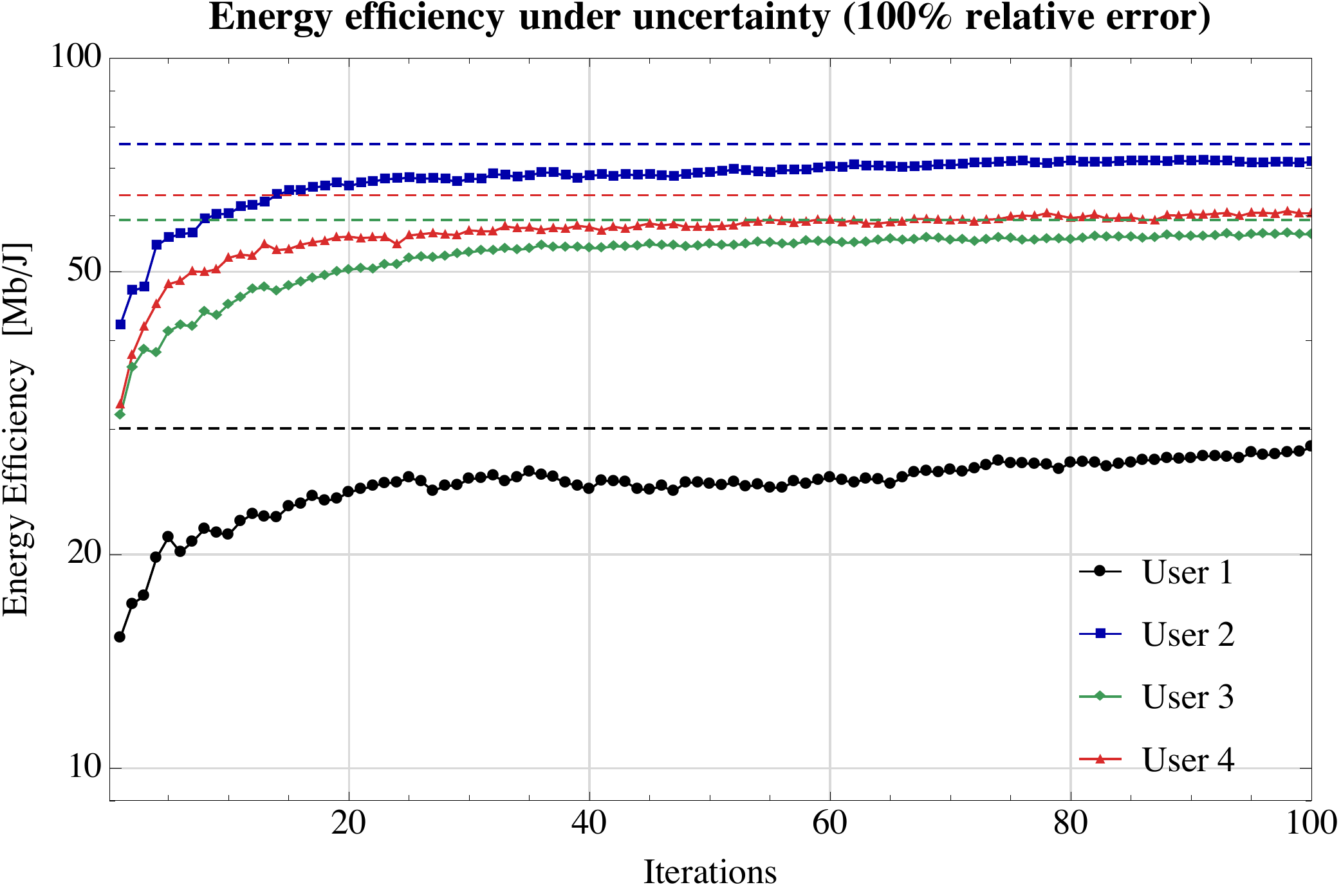}
\\
\caption{Energy efficiency of an online gradient policy with imperfect measurements and observation errors \cite{MB16}.
Even under very high uncertainty, when the gradient estimation lies within $100 \%$ of its true value, the system converges within a few tens of iterations to a stable, no-regret state (dashed line).}
\label{fig:MIMO-noisy}
\end{figure*}


To asses the impact of the noisy estimation in a concrete example, we plot the energy efficiency of the online gradient algorithm in \cite{MB16} in \cref{fig:OGD-MIMO-EE} for four arbitrarily chosen users (see also the corresponding discussion in \cref{sec:OGD}).
Here, the channels are assumed to remain static to better isolate this effect.%
\footnote{In more detail, we consider in \cref{fig:OGD-MIMO-EE} a massive \ac{MIMO} setting with $\tx=8$ transmit antennas and $\rx=128$ receive antennas.
Aside from this and the static channels assumption, all other networks parameters are as in \cref{fig:OGD-MIMO-EE};
complete details can be found in \cite{MB16}.}
Although the system's rate of equilibration is affected by the intensity of the noise, the system still equilibrates within a few tens of iterations, even in the presence of very high uncertainty.
Other applications to rate maximization problems via \ac{OMD} can be found in \cite{MB14, MM16}.

\subsection{Zeroth-order feedback}
\label{sec:zeroth}

The noisy feedback case can be seen as the precursor to a much more challenging question:
\hilite{can the optimizer attain a no-regret state with no gradient feedback whatsoever?}

This question is sometimes referred to as ``\emph{bandit online optimization}'' (in reference to the \acl{MAB} problem)
or
``\emph{online optimization with zeroth-order feedback}'' (in contrast to \emph{first-order} feedback, \ie gradient information).
The motivation for it is clear:
in many practical applications (such as multi-user communication systems), gradient calculations are completely beyond the optimizer's reach (noisy or otherwise), especially if the loss function itself is not revealed.
In that case, the only feedback available to the agent is the actual incurred loss, and the agent must choose a descent direction to follow based only on this minimal piece of information.

In the \acl{MAB} problem, this is achieved by means of the importance sampling technique we described in \cref{sec:bandits} (\cf \cref{alg:EW-partial}).
However, in more general online optimization problems, importance sampling cannot be applied because there is a continuum of points to sample, and also because the value of a function at a given point (a scalar) carries no information about its gradient (a vector).
Thus, to gain such information, it is important to inject some ``directionality'' that could generate a vector out of a scalar.
This is precisely the key idea behind the so-called \acdef{SSA} method of Spall \cite{Spa97} where the gradient is estimated by artificially sampling the function not at the point of interest, but at a nearby, randomly chosen point.

To illustrate the method, it is convenient to begin with the one-dimensional case, \ie functions defined over the real line.
Specifically, suppose that we seek to estimate the derivative of a function $\obj\from\R\to\R$ at some target point $\pivot$ using a single evaluation thereof.
By definition, this derivative can be approximated by the difference quotient
\begin{equation}
\label{eq:difference}
\obj'(\pivot)
	\approx \frac{\obj(\pivot+\mix) - \obj(\pivot-\mix)}{2\mix},
\end{equation}
with the approximation becoming exact as $\mix\to0^{+}$.
Of course, this estimate requires two function evaluations (at $\pivot+\mix$ and $\pivot-\mix$), but it also suggests the following approach:
simply make a (uniform) random draw from $\{\pm1\}$ and sample $\obj$ at $\pivot\pm\mix$ accordingly.
Letting $\unitvec\in\{\pm1\}$ denote the outcome of this draw, the difference quotient above can be rewritten as
\begin{equation}
\frac{\obj(\pivot+\mix) - \obj(\pivot-\mix)}{2\mix}
	= \frac{1}{\mix} \bracks*{\frac{1}{2} \obj(\pivot+\mix) - \frac{1}{2} \obj(\pivot-\mix)}
	= \frac{1}{\mix} \exof{\obj(\pivot+\mix\unitvec) \, \unitvec}.
\end{equation}
\ie as the expectation of the random variable $\obj(\pivot+\mix\unitvec) \unitvec$.
Thus, going back to \eqref{eq:difference}, we see that $\obj'(\pivot)$ can be approximated to $\bigoh(\mix)$ by the single-shot estimator $\mix^{-1} \obj(\pivot+\mix\unitvec) \, \unitvec$.

Extending this construction to functions defined on $\R^{\vdim}$ gives rise to the estimator
\begin{equation}
\label{eq:zeroth}
\frac{\vdim}{\mix} \obj(\pivot + \mix \unitvec) \, \unitvec,
\end{equation}
where $\unitvec$ is now drawn uniformly from the unit sphere $\sphere^{\vdim}$
and
the factor $\vdim$ has been included for dimensional scaling reasons.%
\footnote{Heuristically, the reason for this is that there are $\vdim$ independent directions to sample, each with ``probability'' $1/\vdim$.
Formally, this is a consequence of Stokes' theorem, a fundamental result in differential geometry which is used to prove the estimator's validity.}
As in the one-dimensional case, the bias of the estimator vanishes in the limit $\mix\to0^{+}$:
more precisely, for positive $\mix>0$ and $\Lip$-Lipschitz $\obj$, we have
\begin{equation}
\exof*{\frac{\vdim}{\mix} \obj(\pivot + \mix \unitvec) \, \unitvec}
	= \nabla\obj_{\mix}(\pivot),
\end{equation}
where $\obj_{\mix}$ is a ``$\mix$-smoothed'' approximation of $\obj$ with $\abs{\obj_{\mix}(\pivot) - \obj(\pivot)} \leq \Lip\mix$.%
\footnote{Specifically, $\obj_{\mix}(\act) \equiv \vol(\ball_{\mix})^{-1}\int_{\ball_{\mix}} \obj(\act+\unitvec) \dd\unitvec$ is simply the average value of $\obj$ over $\ball_{\mix}$, a ball of radius $\mix$ centered at $\act$.}
In other words, the single-shot estimator \eqref{eq:zeroth} provides a reasonable approximation of $\nabla\obj(\pivot)$, up to a bias of at most $\bigoh(\mix)$.


\begin{algorithm}[tbp]
\caption{\acf{OGD-0}}
\label{alg:OGD-0}
\input{Algorithms/OGD-0}
\end{algorithm}



\begin{figure}[tbp]
\centering
\input{Figures/OGD-0.tex}
\caption{Schematic representation of \ac{OGD-0} (\cref{alg:OGD-0}).
To ensure feasibility, the projected iterate is mapped to the $\mix$-shrunk region $\feas_{\mix}$.}
\label{fig:OGD-0}
\end{figure}
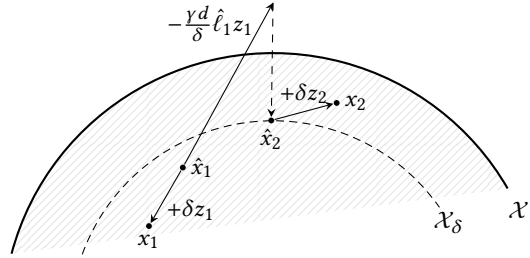


The above is the cornerstone of the ``gradient descent without a gradient'' methodology of \cite{FKM05}.
In particular, the main idea of this method is to apply a projected descent scheme to generate an online sequence of \emph{pivot points} $\pivot_{\run}$.
These points are \emph{not} played by the optimizer, but they act as base points (``pivots'') to select an action at each stage based on the perturbation model
\begin{equation}
\act_{\run}
	= \pivot_{\run} + \mix \unitvec_{\run},
\end{equation}
with $\unitvec_{\run}$ drawn \ac{iid} from the unit sphere $\sphere^{\vdim}$ of $\R^{\vdim}$.%
\footnote{We tacitly assume here that $\feas$ is full-dimensional;
if this is not the case, $\unitvec_{\run}$ should be sampled from the unit sphere of the tangent hull of $\feas$ \textendash\ \ie the smallest subspace of $\R^{\vdim}$ containing $\feas$.}
In so doing, the agent can follow the zeroth-order technique outlined above to simultaneously estimate the (negative) gradient of $\loss_{\run}$ at $\pivot_{\run}$ as
\begin{equation}
\label{eq:estimate-partial}
\est_{\run}
	= -\frac{\vdim}{\mix} \loss_{\run}(\act_{\run}) \, \unitvec_{\run}.
\end{equation}

Of course, this is not an estimate of the loss gradient at $\act_{\run}$.
However, since the distance between the pivot and the agent's chosen action is $\bigoh(\mix)$, the difference is not too big:
optimistically, it should be enough to generate a descent step.
With this in mind, we obtain the following recursive scheme for running \acdef{OGD-0}:
\begin{equation}
\label{eq:OGD-0}
\begin{aligned}
\act_{\run}
	&= \pivot_{\run} + \step\unitvec_{\run}
	\\
\est_{\run}
	&= -\frac{\vdim}{\mix} \loss_{\run}(\act_{\run}) \, \unitvec_{\run}
	\\
\pivot_{\run+1}
	&= \Eucl(\pivot_{\run} + \step\est_{\run})
\end{aligned}
\end{equation}
with $\pivot_{1}$ initialized arbitrarily (for a pseudocode implementation, see \cref{alg:OGD-0}).

Before proceeding with the analysis of \ac{OGD-0}, two remarks are in order:
The first concerns the \emph{bias-variance dilemma:}
the bias of the estimator \eqref{eq:estimate-partial} can be made arbitrarily small but, in so doing, the estimator's variance explodes.
In this way, it is impossible to satisfy both branches of the statistical assumption \eqref{eq:estimate}:
the zeroth-order estimator \eqref{eq:estimate-partial} can either have small bias or small variance, but not both.

The second important consideration is that if the pivot point $\pivot_{\run}$ lies too close to the boundary of $\feas$, the chosen action $\act_{\run} = \pivot_{\run} + \mix\unitvec_{\run}$ may lie outside the problem's feasible region.
To account for this feasibility issue, one can keep the method's pivots away from the boundary of $\feas$ by re-projecting them onto a ``$\mix$-shrunk'' sub-region of $\feas$ (for a schematic illustration, see \cref{fig:OGD-0}).
This introduces a further error of order $\bigoh(\mix)$ in the mix, but since all other estimation errors are of the same order, it does not affect the algorithm's qualitative behavior.
To streamline our presentation and keep our notation as light as possible, we will not carry around this adjustment in the definition of \ac{OGD-0} but we will tacitly assume that it always takes place if/when needed.

With all this at hand, it can be shown that \ac{OGD-0} enjoys the following guarantee \cite{FKM05,SS11}:
\begin{theorem}[Worst-case regret of \ac{OGD-0}]
\label{thm:OGD-0}
Against $\Lip$-Lipschitz convex losses, the \ac{OGD-0} algorithm enjoys the worst-case guarantee
\begin{equation}
\label{eq:reg-zeroth}
\bar\reg_{\horizon}
	= \bigoh(\sqrt{\vdim\Lip} \horizon^{3/4}),
\end{equation}
achieved by taking $\step = (2\horizon)^{-1/2} B / [ \vdim(\loss_{\max}/\mix + \Lip)]$ and $\mix = \horizon^{-1/4} \sqrt{\vdim B\loss_{\max} / (3\Lip)}$,
with $\loss_{\max}$ denoting the agent's maximum possible loss \textpar{\ie $\loss_{\run} \leq \loss_{\max}$ for all $\run$} and $B = \max_{\act\in\feas} \norm{\act}$.
\end{theorem}

The complicated expressions for the operational parameters of \ac{OGD-0} illustrate the delicate dependence of the algorithm's performance on balancing exploration versus exploitation.
If the exploration parameter $\mix$ is too large, the algorithm's gradient estimator will gain in terms of precision (variance) but lose in terms of accuracy;
conversely, for small $\mix$, the estimated gradient will be relatively accurate in the mean, but with immense variability.
This is a consequence of the fact that, in general, exploring a nonlinear function is much more difficult than exploring a bandit with a finite number of arms.

The above also explains the difficulty in closing the gap between the $\bigoh(\horizon^{3/4})$ regret minimization rate of \cref{thm:OGD-0} and the $\bigoh(\horizon^{1/2})$ rate that can be achieved when first-order feedback is available (noisy or otherwise).
This gap \emph{has} been closed in several cases, such as if the optimizer can sample each loss function at two points \cite{ADX10}, if the loss functions are highly smooth \cite{BP16}, and other refinements of the problem.
However, the case of general convex functions remains open and it is not known whether there is a fundamental gap between what is known and what is attainable with zeroth-order feedback (though it is widely believed that $\bigoh(\sqrt{\horizon})$ regret should still be attainable in this case).

\begin{remark*}
Before ending the limited feedback discussion, we note that similar results to \cref{thm:OGD-noisy,thm:OGD-0} also hold for general \ac{OMD} algorithms.
The worse-case regret with noisy feedback is of the same order as in the perfect information case, while zeroth-order feedback incurs the same drop in the rate of regret minimization.
\end{remark*}

\subsection{Fixed horizon vs. anytime bounds}
\label{sec:anytime}

A further consideration regarding the information available to the optimizer is the  knowledge \textendash\ or not \textendash\ of the horizon of play.
Indeed, all the worst-case guarantees that we have discussed so far presume that $\horizon$ is known in advance, so the various parameters of an online optimization algorithm can be fine-tuned as a function thereof.
Depending on the application at hand, this assumption could be severely limiting:
for instance, in the context of telecommunication networks, it is often far-fetched to assume that a wireless user can know in advance the duration of their call.

Guarantees established for problems where the duration of play is unknown are usually referred to as \emph{anytime bounds} and one of the main techniques to achieve them is via the so-called \emph{doubling trick} \cite{CBL06}.
The basic idea behind this device is as follows:
the optimizer chooses a fixed window consisting of $\window$ rounds and plays with a no-regret algorithm optimized for $\window$ stages.
When this window draws to a close, the agent doubles the window length ($\window \leftarrow 2\window$) and restarts the algorithm, now optimized for $2\window$ stages.
This process then repeats as needed, until play ends \textendash\ specifically, after no more than $\ceil{\log_{2}(\horizon/\window)}$ period doublings.

In so doing, if the chosen algorithm enjoys an upper bound of $M\window^{1/2}$ when run over a fixed window of size $\window$, the induced anytime bound will be
\begin{equation}
\smashoperator{\sum_{k=0}^{\ceil{\log_{2}(\horizon/\window)}}} M \sqrt{2^{k}\window}
	= M\sqrt{\window} \frac{\sqrt{2}^{\ceil{\log_{2}(\horizon/\window)}+1}-1}{\sqrt{2} - 1}
	\leq M\sqrt{\window} \, \frac{2 \sqrt{\horizon/\window}}{\sqrt{2} - 1}
	= \frac{2}{\sqrt{2} - 1} M\sqrt{\horizon}.
\end{equation}
In other words, thanks to the doubling trick, the optimizer can achieve an anytime regret guarantee which is at most a factor of $2/(\sqrt{2} - 1)$ worse than the algorithm's fixed horizon guarantee.
Moreover, this constant factor does not depend on the size of the window, so the optimizer has significant flexibility in applying the doubling trick.

Similar tricks can also be applied to the case where an algorithm guarantees a regret of $M\window^{\alpha}$ for some $\alpha\in(0,1)$.
In this case, it is again possible to achieve $\bigoh(M\window^{\alpha})$ regret up to a universal multiplicative factor that depends only on $\alpha$.
We leave the details as an exercise for the reader.

\subsection{Dynamic regret}
\label{sec:dynamic}

To conclude this section, we briefly return to the definition of the regret as the performance gap between the optimizer's chosen policy and the best \emph{fixed} action in hindsight.
This definition is somewhat restrictive as it only compares the chosen policy to \emph{constant} policies;
ideally, the optimizer would seek to minimize the incurred \emph{dynamic regret}, defined here as
\begin{equation}
\label{eq:reg-dynamic}
\dreg_{\horizon}
	= \sum_{\run=1}^{\horizon} \bracks*{\loss_{\run}(\act_{\run}) - \min_{\act\in\feas} \loss_{\run}(\act)}.
\end{equation}
Obviously, since $\dreg_{\horizon}$ targets the smallest possible aggregate loss $\sum_{\run=1}^{\horizon} \min_{\act\in\feas} \loss_{\run}(\act)$, we have $\reg_{\horizon} \leq \dreg_{\horizon}$, \ie the ordinary (static) regret is always less than or equal to the dynamic regret.
Unfortunately for the optimizer, this inequality indicates an insurmountable gap:
the sequence of payoff vectors used to establish Cover's impossibility result in \cref{sec:bandits} can be easily adapted to show that the adversary can always enforce $\dreg_{\horizon} = \Omega(\horizon)$, even when restricted to linear loss functions.
Thus, in general, it is not possible to achieve sublinear dynamic regret against an informed adversary.

The key limitation here is the variability of the adversary's choices:
recent work by Besbes et al. \cite{BGZ15} showed that it is possible to achieve sublinear dynamic regret if the so-called ``variation budget''
\begin{equation}
\label{eq:budget}
\mathrm{VB}_{\horizon}
	= \sum_{\run=1}^{\horizon-1} \norm{\loss_{\run} - \loss_{\run+1}}
\end{equation}
of the sequence of loss functions chosen by the adversary is itself sublinear in $\horizon$.
This is achieved by exploiting an algorithm that guarantees sublinear (static) regret and then restarting it following a judiciously chosen schedule (not unlike the doubling trick described above).
Otherwise, if $\mathrm{VB}_{\horizon}$ scales as $\Theta(\horizon)$, minimizing the agent's dynamic regret is a bridge too far.

Other, less ``dynamic'' notions of regret have also been studied in the literature (such as the ``adaptive'' or ``shifting'' regret).
For a discussion of the guarantees available in that context, we refer the reader to \cite{CBL06}.


%% file: Algorithms/OGD-0.tex

\begin{algorithmic}[1]
\REQUIRE
	step-size $\step>0$, exploration factor $\mix>0$
\STATE
	choose pivot point $\pivot$
	\COMMENT{initialization}%
\FOR{$\run=1$ \TO $\horizon$}
	\STATE	
		draw $\unitvec$ uniformly from $\sphere^{\vdim}$
		\COMMENT{exploration direction}%
	\STATE
		play $\act \leftarrow \pivot + \mix\unitvec$
		\COMMENT{choose action}%
	\STATE
		incur loss $\hat\loss \leftarrow \loss_{\run}(\act)$
		\COMMENT{losses revealed}%
	\STATE
		set $\est \leftarrow -\frac{\vdim}{\mix} \hat\loss \cdot \unitvec$
		\COMMENT{estimate gradient}%
	\STATE
		set $\pivot \leftarrow \Eucl(\pivot + \step\est)$
		\COMMENT{update pivot}%
\ENDFOR
\end{algorithmic}

%% file: Figures/OGD-0.tex

\colorlet{TangentColor}{blue}
\colorlet{PolarColor}{red}

\begin{tikzpicture}
[scale=1.2,
nodestyle/.style = {circle,fill=black,inner sep=0, minimum size=2}]

\small

\draw [thick,pattern = north east lines, pattern color = black!10] (165:3) arc (165:30:3);
\draw node at (25:3) {$\feas$};
\draw [densely dashed] (160:2.25) arc (160:35:2.25);
\draw node at (30:2.25) {$\feas_{\mix}$};

\node [nodestyle] (pivot) at (120:2) {.};
\node [right] at (pivot) {$\pivot_{1}$};

\node [nodestyle] (action) at ($(pivot)+(-120:.75)$) {.};
\node [below] at (action) {$\act_{1}$};

\node (estimate) at ($(pivot)+(60:2)$) {};

\draw[-stealth] (pivot) -- (action) node [near end, right]{$+\mix\unitvec_{1}$};
\draw[-stealth] (pivot) -- (estimate.north) node [very near end, left]{$-\frac{\step\vdim}{\mix} \hat\loss_{1} \unitvec_{1}$};

\node [nodestyle] (pivot2) at ($(pivot)+(28:1.11)$) {.};
\node [below] at (pivot2) {$\pivot_{2}$};
\draw[-stealth,dashed] (estimate.north) -- (pivot2) node [midway, right]{};

\node [nodestyle] (action2) at ($(pivot2)+(15:.75)$) {.};
\node [right] at (action2) {$\act_{2}$};
\draw[-stealth] (pivot2) -- (action2) node [midway, above]{$+\mix\unitvec_{2}$};

\end{tikzpicture}

%% file: Conclusions.tex

The aim of this tutorial was to provide an introduction to online optimization and online learning with a clear focus on their applications to signal processing \textendash\ from resource allocation problems in wireless communication networks to multimedia indexing and, in particular, metric learning for image similarity search and data classification.
A major asset of this framework is its flexibility and the fact that it already encompasses
discrete choice models,
offline optimization problems (both static and stochastic),
but also non-stationary problems in which the objective's dynamics might be completely unpredictable.
Another advantage lies in the theoretical guarantees of the developed algorithms which are able to reach the best average performance of an ideal policy with full knowledge of the future, despite being run with minimal coordination and feedback assumptions.
We find this property to be one of the most appealing characteristics of online optimization, serving to explain its wide applicability in large-scale machine learning and Big Data problems.